\documentclass[conference]{IEEEtran}
\usepackage{tikz}
\usepackage{amsmath}
\usepackage{url}
\usepackage{hyperref}
\usepackage{amsfonts}

\usepackage{xcolor}
\definecolor{ForestGreen}{RGB}{34, 139, 34}
\usepackage{tabularx}
\usepackage{lipsum}
\usepackage{makecell}
\usepackage{tcolorbox}

\usepackage{booktabs} 
\usepackage{multirow} 
\usepackage{graphicx} 
\usepackage{algorithm}
\usepackage{algpseudocode}
\usepackage{soul}

\ifCLASSINFOpdf
\else
\fi
\begin{document}
%
\title{
\texttt{ViGText}: Deepfake Image Detection with Vision-Language Model Explanations and Graph Neural Networks}

\author{
\IEEEauthorblockN{
Ahmad ALBarqawi\IEEEauthorrefmark{1},
Mahmoud Nazzal\IEEEauthorrefmark{4},
Issa Khalil\IEEEauthorrefmark{6},
Abdallah Khreishah\IEEEauthorrefmark{1}, and
NhatHai Phan\IEEEauthorrefmark{1}
}
\IEEEauthorblockA{\IEEEauthorrefmark{1}New Jersey Institute of Technology, Newark, NJ, USA\\}
\IEEEauthorblockA{\IEEEauthorrefmark{4}Old Dominion University, Norfolk, VA, USA\\}
\IEEEauthorblockA{\IEEEauthorrefmark{6}Qatar Computing Research Institute (QCRI), HBKU, Doha, Qatar\\
Emails: aka87@njit.edu, mnazzal@odu.edu, ikhalil@hbku.edu.qa, \{abdallah, phan\}@njit.edu
}
}


%


\IEEEoverridecommandlockouts
\makeatletter\def\@IEEEpubidpullup{6.5\baselineskip}\makeatother
\IEEEpubid{\parbox{\columnwidth}{
		Network and Distributed System Security (NDSS) Symposium 2026\\
		23 - 27 February 2026 , San Diego, CA, USA\\
		ISBN 979-8-9919276-8-0\\
		https://dx.doi.org/10.14722/ndss.2026.230303\\
		www.ndss-symposium.org
}
\hspace{\columnsep}\makebox[\columnwidth]{}}

\maketitle

\begin{abstract}
The rapid rise of deepfake technology, which produces realistic but fraudulent digital content, threatens the authenticity of media. Deepfakes manipulate videos, images, and audio, spread misinformation, blur the line between real and fake, and highlight the need for effective detection approaches. Traditional deepfake detection approaches often struggle with sophisticated, customized deepfakes, especially in terms of generalization and robustness against malicious attacks. This paper introduces \texttt{ViGText}, a novel approach that integrates images with \underline{Vi}sion Large Language Model (VLLM) \underline{Text} explanations within a \underline{G}raph-based framework to improve deepfake detection. The novelty of \texttt{ViGText} lies in its integration of detailed explanations with visual data, as it provides a more context-aware analysis than captions, which often lack specificity and fail to reveal subtle inconsistencies. \texttt{ViGText} systematically divides images into patches, constructs image and text graphs, and integrates them for analysis using Graph Neural Networks (GNNs) to identify deepfakes. Through the use of multi-level feature extraction across spatial and frequency domains, \texttt{ViGText} captures details that enhance its robustness and accuracy to detect sophisticated deepfakes. Extensive experiments demonstrate that \texttt{ViGText} significantly enhances generalization and achieves a notable performance boost when it detects user-customized deepfakes. Specifically, average F1 scores rise from 72.45\% to 98.32\% under generalization evaluation, and reflects the model’s superior ability to generalize to unseen, fine-tuned variations of stable diffusion models. As for robustness, \texttt{ViGText} achieves an increase of 11.1\% in recall compared to other deepfake detection approaches against state-of-the-art foundation model-based adversarial attacks. \texttt{ViGText} limits classification performance degradation to less than 4\% when it faces targeted attacks that exploit its graph-based architecture and marginally increases the execution cost. \texttt{ViGText} combines granular visual analysis with textual interpretation, establishes a new benchmark for deepfake detection, and provides a more reliable framework to preserve media authenticity and information integrity.\end{abstract}
\section{Introduction}
Recent advancements in deep learning, particularly in generative models, have enabled the creation of highly realistic synthetic media. The term \textit{deepfake} refers to synthetic content generated by altering or replacing a person’s appearance or voice in images, videos, or audio, which makes it increasingly difficult to distinguish from authentic media \cite{westerlund2019emergence}.  The rise of deepfake technology introduces serious challenges to the accuracy and trustworthiness of digital media, and raises concerns in domains such as politics, media, and entertainment \cite{vaccari2020deepfakes}. Along this line, recent reports have highlighted a surge in deepfake pornography targeting young women, including underage individuals, in South Korea \cite{key,yahooSouthKorea}. Alarmingly, the Korean Teachers Union reports that over 200 schools have been impacted, with a notable increase in deepfakes which target teachers in recent years, according to the Ministry of Education \cite{bbcSouthKorea}. In Ukraine, deepfakes have been used to disseminate misinformation and manipulate public perception during the ongoing conflict \cite{theconversationDeepfakesWarfare,nytimesDeepfakeUS}. This AI-generated content, which convincingly fabricates events or statements, contributes to public confusion and complicates the distinction between truth and deception. As deepfakes become more advanced, the ability to distinguish between real and synthetic content becomes increasingly difficult. As the prevalence of deepfake videos and images grows, it not only fuels the spread of misinformation but also poses significant threats to privacy \cite{kugler2021deepfake}, security \cite{pantserev2020malicious}, and public trust \cite{pawelec2022deepfakes}. These concerns have driven extensive research which aim to detect deepfakes, and emphasize the urgent need for effective solutions to address this escalating issue \cite{Tolosana2020,Afchar2018,Nguyen2019,Verdoliva2020}.

\begin{figure}[!thb]
\centering
\includegraphics[width=0.9\columnwidth]{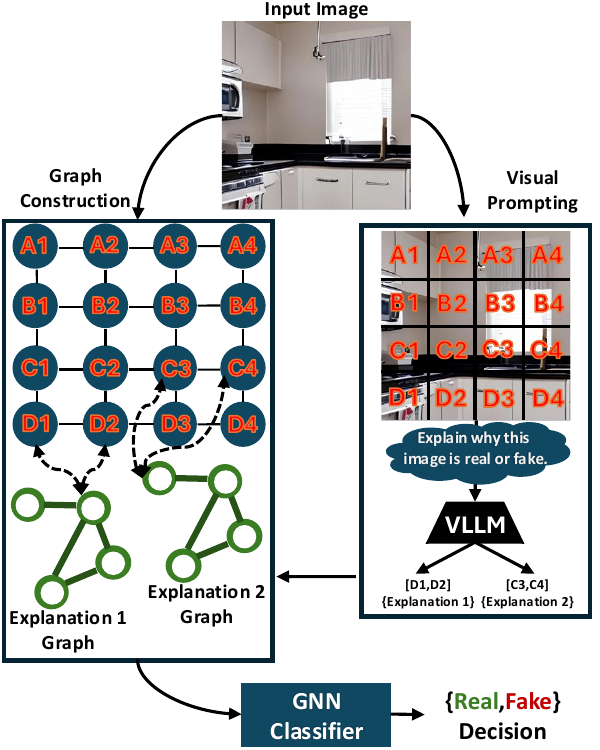}
\vspace{-0.2cm}
\caption{An overview of the proposed \texttt{ViGText}'s approach.}
\label{approach}
\end{figure}

\textbf{Prior Approaches and Limitations.} Recent efforts in deepfake image detection have mainly relied on learning-based methods. These approaches typically start with a labeled dataset that contains real and deepfake images. The objective is to train a model to detect deepfakes and generalize this detection capability to new, unforeseen deepfakes. Commonly used models include Convolutional Neural Networks (CNNs) and simpler, more traditional models such as feed forward neural networks. \cite{chai2020makes} uses CNNs to classify image patches, while \cite{liu2020global} enhances detection by capturing global textures with \textit{Gram-Net}. \cite{he2021beyond} focuses on re-synthesizing images to analyze residual errors, and \cite{wang2020cnn} demonstrates cross-architecture generalization in CNN classifiers. Additionally, \cite{afchar2018mesonet} presents lightweight CNNs that target mesoscopic image properties. More recently, there has been a shift towards simpler models and novel feature extraction techniques. \cite{ricker2022towards} utilizes frequency domain features with a basic feed-forward neural network, \cite{sha2023fake} combines image and caption features from a foundation model, and \cite{ojha2023towards} shows that a linear layer trained on features from a large and non-deepfake-specific foundation model can achieve effective classification and generalization. Sifat et al. in \cite{abdullah2024analysis} analyze current deepfake detection methods and identify key challenges. They demonstrate that many existing approaches fail to generalize effectively when exposed to user-customized or fine-tuned versions of the generative models used during training. Additionally, they highlight that these methods are highly susceptible to adversarial attacks generated with advanced foundation models. These findings point to a significant gap in current techniques, which struggle to adapt to the unpredictable nature of deepfakes in real-world scenarios.

\par \textbf{Challenges.} Deepfake detection faces solid challenges, particularly in achieving robustness against an ever-evolving threat landscape. The rapid development of generative AI technologies frequently outpaces detection capabilities, which leaves defenses unprepared to address new attack vectors \cite{abdullah2024analysis}. Another key challenge lies in the integration of textual and visual data. Current approaches rely on image captions that lack the depth and specificity needed to capture the subtle characteristics required for effective detection \cite{sha2023fake}. Even in scenarios where detailed textual data is available, it is non-trivial to integrate this information with visual data in a meaningful way \cite{sha2023fake}. Traditional methods often concatenate visual and textual embeddings in a straightforward manner, that fails to fully utilize their complementary properties. Another challenge is the need for generalizability across a wide variety of deepfake models. Existing methods frequently exhibit significant performance degradation when applied to fine-tuned or customized generative models, which emphasizes their lack of adaptability to novel threats \cite{abdullah2024analysis}. To overcome these challenges, it is critical to explore advanced integration strategies that can effectively combine textual and visual information, and integrate their complementary strengths.

\par \textbf{An Overview of the Proposed \texttt{ViGText}.} To overcome the above challenges, we present \texttt{ViGText}, a new deepfake detection approach that integrates image analysis and text-based explanations from a Vision Large Language Model (VLLM) in a graph-based framework, as shown in Fig. \ref{approach}. The novelty of \texttt{ViGText} lies in its ability to unify visual and textual analysis through tailored graph construction, which allows for the detection of subtle inconsistencies with enhanced generalization and robustness. The process begins as the input image is divided into square patches, each represented as a node in an \textit{image graph}. Each patch is embedded while taking into account spatial and frequency information, with the latter extracted using the Discrete Cosine Transform (DCT). Edges are then added to connect adjacent patches, to capture local spatial dependencies. Alongside the image graph, an \textit{explanation graph} is constructed with the use of a VLLM to generate textual explanations for the patches. Explanation graphs are integrated with the image graph such that each explanation is connected to the patches it describes, which forms a dual-graph structure. This dual graph is then analyzed by a Graph Neural Network (GNN) to determine whether the image is real or fake. While recent methods use image captions for deepfake detection \cite{sha2023fake}, \texttt{ViGText} integrates detailed textual explanations within a graph structure, combining spatial and frequency features to improve generalization to fine-tuned models and robustness against adversarial images.

\par \textbf{Summary of Contributions.} This work presents the following contributions.
\begin{list}{$\bullet$}{\leftmargin=0em \itemindent=0em}
\item Introduction of a Dual-Graph Framework for Enhanced Detection: We propose \texttt{ViGText}, a novel approach that unifies image analysis with textual explanations generated by VLLMs. \texttt{ViGText} embeds each image patch as it considers both spatial and frequency features, then organizes the visual and textual data into a dual-graph structure which achieves a more robust integration for deepfake detection.
\item Enhanced Generalization to Diverse Generative Models: \texttt{ViGText} achieves superior generalization across a wide variety of user-customized, fine-tuned variants of generative models without the need to train on the base model images. Through effective integration of context-aware explanations and frequency-domain features in the graph framework, \texttt{ViGText} mitigates the substantial performance degradation observed in prior methods when confronted with user-customized models.
\item Robustness Against Evolving Threats: \texttt{ViGText} demonstrates strong resilience against adversarial attacks, that include novel foundation model-based threats. This robustness addresses vulnerabilities where existing methods fail against adversarial manipulations intentionally crafted through advanced vision foundation models.
\item Extended Testing on Generalization Datasets: To evaluate generalization in diverse scenarios, we introduce an extended dataset that comprises eight new testing sets derived from user-customized fine-tuned variants of the Stable Diffusion 3.5 model \cite{AI_2024}. This expansion, alongside pre-existing datasets, enables a more comprehensive assessment of detection performance across a broader range of generative models.

\end{list}
\section{Background}

\textbf{Deepfakes.} Deepfakes are a form of synthetic media where artificial intelligence (AI) is used to create hyper-realistic but fake images, videos, or audio recordings. The technology behind deepfakes involves advanced machine learning techniques, particularly deep learning algorithms, which analyze large datasets of real images or audio to generate new, highly convincing content \cite{korshunova2017fast,zhang2021flow}. Initially developed for entertainment and creative purposes, deepfakes have rapidly evolved, raising significant ethical, legal, and societal concerns \cite{meskys2020regulating,gamage2022deepfakes}. Deepfakes can be used maliciously to fabricate videos of individuals saying or doing things they never actually did, leading to potential harm, such as misinformation, identity theft, and reputational damage \cite{mustak2023deepfakes,collins2019forged}. The growing accessibility of deepfake technology has sparked global debates on the need for regulation, detection methods, and public awareness to mitigate the risks associated with this powerful technology.

\par \textbf{Graph Neural Networks.} Graph neural networks (GNNs) \cite{kipf2016semi} have significantly advanced the field of deep learning by extending it to graph-structured data. These models process messages across graph edges and aggregate this information at nodes. The workflow of a GNN entails extracting low-dimenstional embeddings from a graph inputs utilizing boith local node features and grpah topology. GNNs are known to be effective as classifiers in various domains. For instance, in fraud detection \cite{wu2020comprehensive}, they utilize the relational information in transaction networks. Similarly, in drug discovery \cite{zhou2020graph}, GNNs help in understanding molecular structures. They have also been successfully applied in social network analysis \cite{fan2019graph} and recommendation systems \cite{ying2018graph}, demonstrating their versatility and strength as classifiers. The ability of GNNs to handle complex relational data and their adaptability to different types of graph-structured information make them a powerful tool in numerous state of the art (SoTA) applications.

\par \textbf{NLP Using GNNs} GNNs have emerged as a powerful tool in Natural Language Processing (NLP) by effectively capturing complex dependencies within text through graph representations. Unlike traditional sequence-based models, GNNs enable the modeling of syntactic and semantic relationships by representing words, sentences, or documents as nodes in a graph, with edges characterizing various linguistic connections \cite{wu2023graph,li2019recursive}. This approach has proven particularly effective in tasks such as relation extraction \cite{zhang2018graph}, text classification \cite{yao2019graph,wu2023graph}, and sentiment analysis \cite{singh2020sentiment}, where understanding the intrinsic structure of language is crucial.

\par \textbf{Vision Large Language Models and Visual Prompting} Vision Large Language Models (VLLMs) represent a rapidly advancing area of artificial intelligence that combines visual and textual data to perform a wide range of tasks, including image captioning \cite{chen2022visualgpt,zhou2020unified}, question answering \cite{zhou2020unified}, classification, and segmentation \cite{xu2022simple}. A notable recent development in this field is the concept of visual prompting \cite{yang2024fine,lin2024draw}. This technique involves using specific visual instructions to guide a model’s interpretation or generation of text, akin to how text prompting is used in natural language processing to generate responses based on textual prompts. In visual prompting, the model is given an image or a modified version of an image, rather than relying solely on textual input. Recent research has demonstrated the effectiveness of visual prompting in improving the adaptability and performance of AI models across various tasks. By incorporating visual instructions such as marks or annotations, these techniques assist models in tackling complex tasks, such as robotic manipulation \cite{liu2024moka}, image processing \cite{shi2024dragdiffusion}, and perception \cite{yang2023set}, without requiring additional fine-tuning.

\begin{figure}[!thb]
\resizebox{0.99\columnwidth}{!}{
\includegraphics[width=12cm]{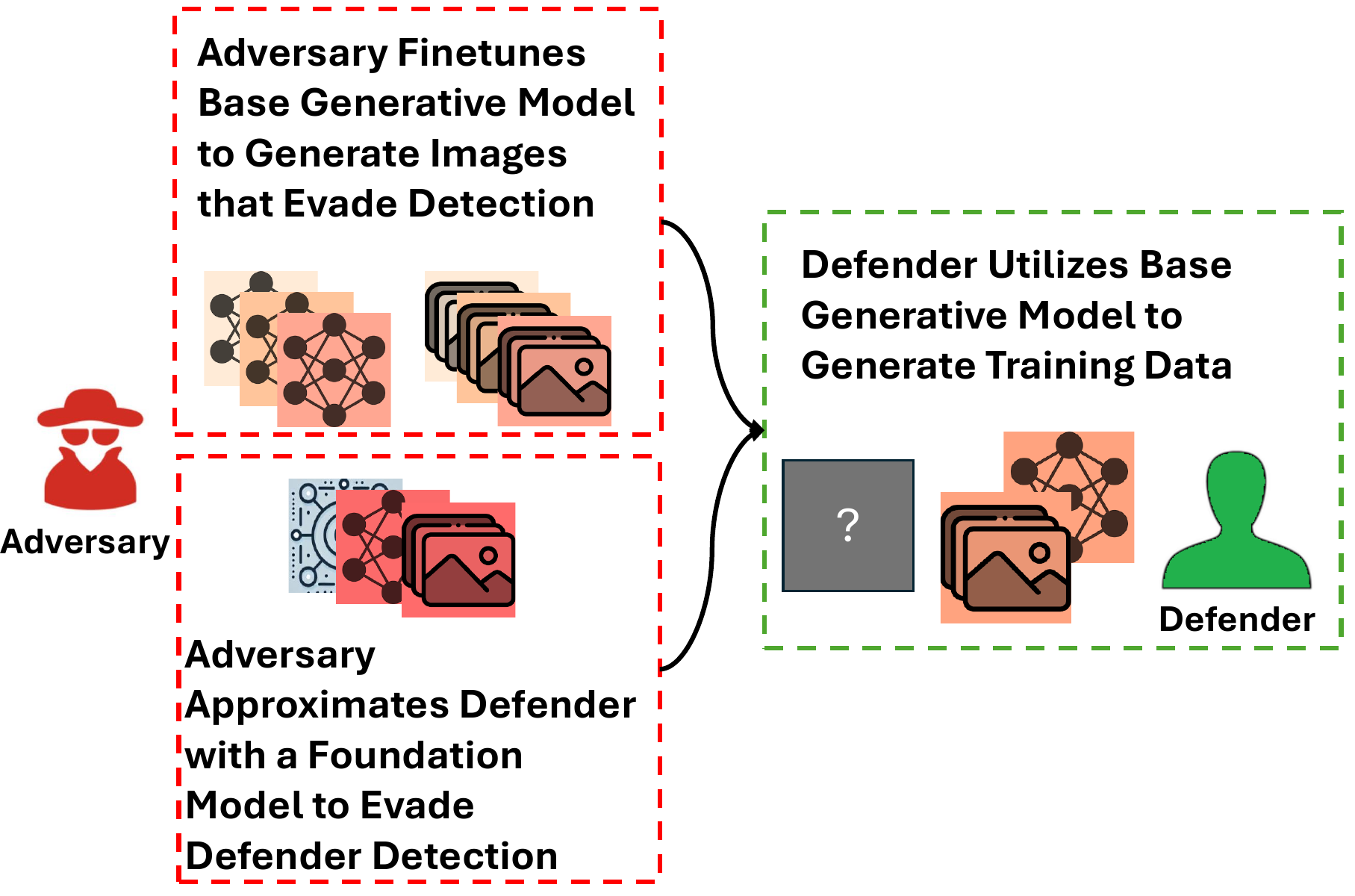}
}
\vspace{-0.1cm}
\caption{An Illustration of the threat model.}
\label{threat}
\end{figure}

\section{Threat Model and Assumptions}

The threat model includes an adversary (malicious actor) who generates deepfakes and targets detection evasion, while a defender tries to detect these deepfakes. This interaction is depicted in Fig. \ref{threat}. We characterize the threat model by detailing the objectives, knowledge, and capabilities of both the adversary and the defender. The adversary’s objective is to produce deepfakes that evade detection by the defender’s system. The adversary possesses advanced generative technologies, including the ability to fine-tune base models to create diverse and realistic deepfakes. Recent parameter-efficient fine-tuning techniques, such as Low-Rank Adaptation (LoRA) \cite{hu2021lora}, have made it feasible for adversaries with limited computational resources to create customized variants of foundational models like Stable Diffusion \cite{podell2023sdxl}. These fine-tuned variants introduce subtle variations that make detection challenging and degrade performance even without explicit knowledge of the defender’s methods \cite{abdullah2024analysis}. Moreover, while our experiments involve edits made using StyleCLIP which is a text-driven image editing method operating in the latent space of StyleGAN2 \cite{karras2020analyzing}, we do not claim these represent real-world partial manipulations such as face swaps or reenactments. Instead, we use these edits to simulate a distinct adversarial behavior: minimal and localized semantic changes that retain the identity and overall image context. This falls within the broader category of fully synthetic images but provides a useful starting point for studying targeted manipulations that aim to evade detection through subtle changes rather than drastic ones. These controlled manipulations help us study the challenge of detecting small but meaningful changes, adding a different kind of adversarial tactic to our threat model.
\par As for the defender, its objective is to efficiently detect deepfakes and generalize detection capabilities across a wide range of fine-tuned variants. Access to the base generative model, from which adversaries derive their fine-tuned variants, is a practical and effective approach because many generative models are publicly available or widely accessible on website like huggingface \cite{huggingface}. Fine-tuning techniques like LoRA \cite{hu2021lora} typically modify only specific layers or parameters while retaining the core characteristics of the base model. These shared traits, such as architecture patterns, feature representations, and generative tendencies, remain largely intact across variants. By focusing on these foundational traits, the defender can generalize detection capabilities to fine-tuned variants without the impractical need to train on every possible one. Another objective for the defender is to ensure robustness against adversarial attacks, particularly those generated using advanced foundation models. These models enable adversaries to craft subtle, highly deceptive manipulations that can bypass detection. Robustness is critical to maintain the reliability of detection systems in the face of increasingly sophisticated and adaptive threats. For this objective, the defender aims to create systems that can withstand evolving attack strategies while preserving the integrity of their results.
 
\begin{figure}[!htb]
\centering
\resizebox{0.98\columnwidth}{!}{
\includegraphics{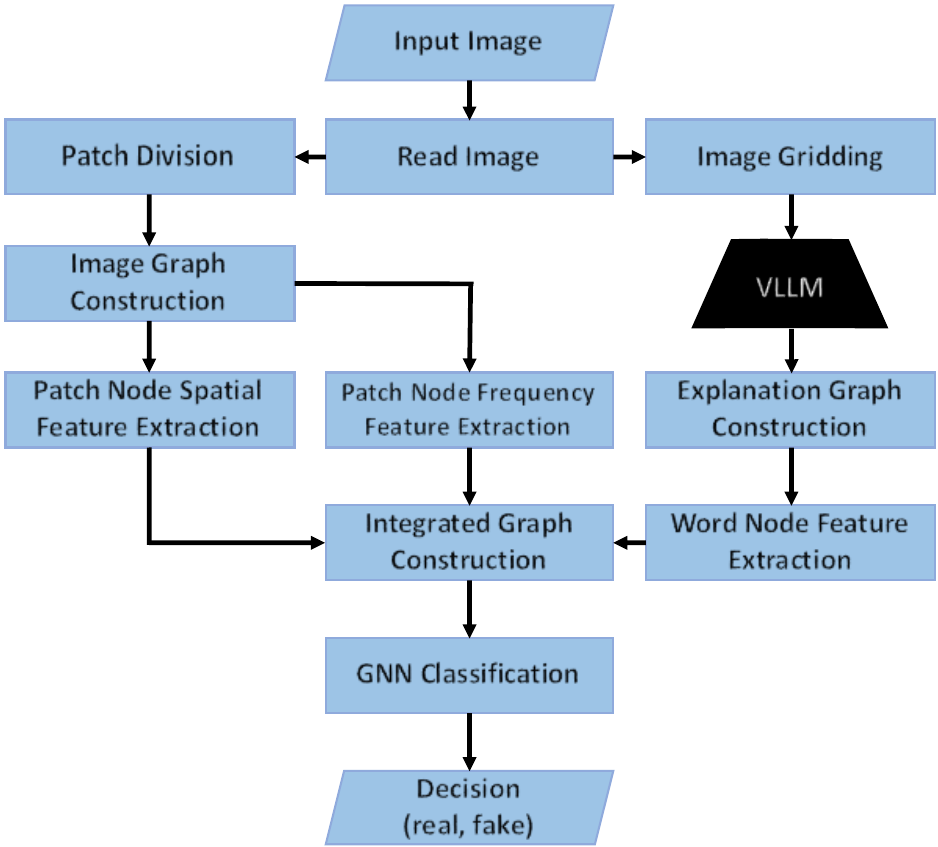}
}
\vspace{-0.1cm}
\caption{Block diagram of the proposed \texttt{ViGText} pipeline, illustrating the key components and processes.}
\label{pipeline}
\end{figure}

\section{The Proposed \texttt{ViGText}}
\label{section4}

\par In this section, we detail the \texttt{ViGText} approach, which is illustrated in the block diagram in Fig. \ref{pipeline} and involves constructing graphs from both image patches and generated explanations, extracting features, and integrating these graphs before utilizing a GNN for detection. We begin by discussing the motivation for using explanations generated by VLLMs. Following this, we formulate the problem of deepfake detection with the availability of images with text. We then introduce a graph-based framework that combines textual and image data, providing a richer context for analysis. Finally, we describe the process of generating these textual explanations and their subsequent integration with images into a unified graph structure.

\subsection{From Captions to Explanations}

\begin{figure}[!htb]
\centering
\resizebox{0.99\columnwidth}{!}{
\includegraphics[width=0.99cm]{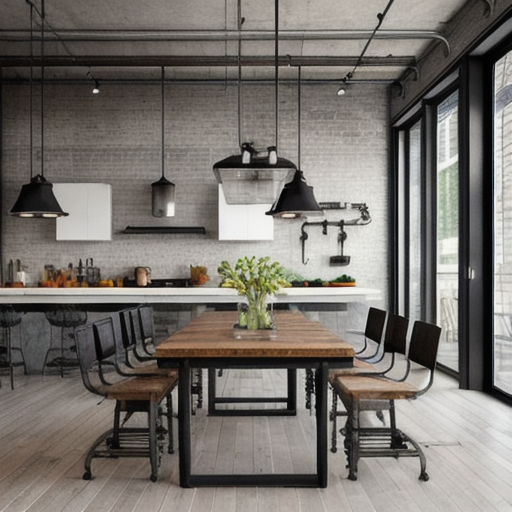}}
\vspace{-0.1cm}
\caption{Generated image misclassified as real by DE-FAKE \cite{sha2023fake}.}
\label{example}
\end{figure}

\par \texttt{ViGText} builds upon a similar concept used in existing techniques like DE-FAKE \cite{sha2023fake}, which incorporates image captions alongside visual data for deepfake detection. However, captions often provide only broad descriptions of the image, which lack the specificity needed to identify inconsistencies. For example, as shown in Fig. \ref{example}, DE-FAKE misclassifies a deepfake as real based on the caption "a kitchen and dining area", which describes the scene in generic terms without addressing visual details that might indicate manipulation. For instance, a VLLM-generated explanation might state, "The cabinets and hanging lights show natural reflections and shadows, indicating a real environment", or "The table and chairs have detailed textures and consistent lighting, which are characteristic of a real image".

\par These detailed explanations capture specific features that contribute to an image’s perceived authenticity. However, VLLMs alone are not capable of accurately classifying images as real or fake \cite{jia2024can}. This is where \texttt{ViGText} excels. {\em Through the analysis of both the visual content and the corresponding explanations, \texttt{ViGText} identifies inconsistencies between the described features and the actual visual elements.} For instance, if an explanation mentions realistic shadows and reflections, but the image lacks these elements or displays unnatural artifacts, such discrepancies serve as strong indicators that the image may be a deepfake. The combination of detailed explanations with visual analysis allows \texttt{ViGText} to address the limitations of caption-based approaches like DE-FAKE and delivers a more robust and reliable framework for deepfake detection.

\subsection{Problem Formulation}

\par With the above-mentioned explanations, the problem of deepfake detection, with a focus on generalizability and adversarial robustness, requires optimizing a classifier function \( f \) that maps an input image \( I \) and its corresponding explanation \( E \) to a binary output \( \{0, 1\} \), where 0 indicates a real image and 1 indicates a fake image. In a machine learning-based solution, the objective is to maximize the accuracy of this classifier over a dataset \( \mathcal{D} = \{(I_i, E_i, y_i)\}_{i=1}^n \), where \( y_i \) are the ground truth labels as expressed as (\ref{eq1}).
\begin{equation}
\begin{alignedat}{2}
& \text{Maximize:} \quad && \frac{1}{n} \sum_{i=1}^n \mathbb{I}(\mathcal{f}(I_i, E_i) = y_i) \\
& \text{Subject to:} && \min_{\delta \in \Delta} \frac{1}{n} \sum_{i=1}^n \mathbb{I}(f(I_i + \delta, E_i) = y_i) \geq \tau_r, \\
& && \mathbb{E}_{(I, E, y) \sim \mathcal{D}_{new}}[\mathbb{I}(f(I, E) = y)] \geq \tau_g,
\label{eq1}
\end{alignedat}
\end{equation}
\noindent where \( I_i \) are the input images, \( E_i \) are the generated explanations, and \( y_i \) are the labels (0 for real, 1 for fake). The indicator function \( \mathbb{I}(\cdot) \) returns 1 if its condition is true and 0 otherwise. \( \Delta \) represents allowable perturbations for testing robustness, with \( \tau_r \) as the required robustness threshold. \( \mathcal{D}_{new} \) denotes the distribution of unseen data, \( \tau_g \) is the generalization threshold that must be met, and \( \mathbb{E} \) denotes expectation. This problem requires careful consideration of how images and textual data are integrated to fully grasp the benefits of including textual information. \( f \) is the binary classifier characterizing detection, \( y \) is the ground truth label (0 for real, 1 for fake).

\subsection{Visual and Textual Integration}

\par The integration of textual and visual data is crucial for effective deepfake detection. While DE-FAKE \cite{sha2023fake} uses simple concatenation of embeddings for captions and images, this approach fails to capture detailed interdependencies. In contrast, \texttt{ViGText} employs a graph-based model, which integrates textual explanations and visual data to establish meaningful relationships.
\par To explore the impact of integration methods, the following experiment is conducted, we compare two approaches: DE-FAKE, which uses explanations as textual input but arbitrarily concatenates their embeddings with image embeddings, and \texttt{ViGText}. Table \ref{vsdefake} summarizes the results of this comparison.
\begin{figure}[!htb]
\centering
\resizebox{0.99\columnwidth}{!}{
\begin{tabular}{cc}
\includegraphics[width=10cm]{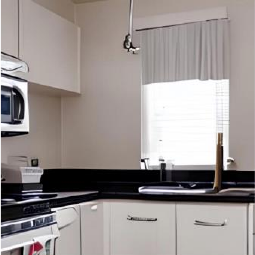}
&
\includegraphics[width=10cm]{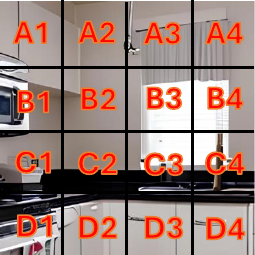}
\\
\Huge{(a)}
&
\Huge{(b)}
\end{tabular}}
\vspace{-0.1cm}
\caption{Image overlay with grid: (a) original, (b) with grid overlay.}
\label{grid} 
\end{figure}

\begin{table}[!h]
\centering
\caption{Performance Comparison of \texttt{ViGText} vs. DE-FAKE \cite{sha2023fake} when both approaches are given the same explanations as the textual input (The highest is in bold).}
\label{vsdefake}
\resizebox{\columnwidth}{!}{ 
\begin{tabular}{|c|c|c|c|c|}
\hline
 & Accuracy & Recall & Precision & F1 \\ \hline
DE-FAKE w/Explanations & 90.00 & 91.20 & 89.00 & 90.10 \\ \hline
\texttt{ViGText} & \textbf{99.25} & \textbf{99.80} & \textbf{98.52} & \textbf{99.26} \\ \hline
\end{tabular}
}
\end{table}

\par The results in Table \ref{vsdefake} show a clear difference in performance. Simply replacing captions with explanations and concatenating their embeddings with image features does not lead to accurate detection. DE-FAKE achieves an accuracy of 90\%, despite the use of richer explanations instead of simple captions. This showcases a key limitation: the arbitrary integration of textual and visual data by concatenation fails to capture their complex interdependencies. Unlike the simple techniques used in DE-FAKE, the graph structure allows \texttt{ViGText} to capture intricate interdependencies between textual explanations and visual features, which enables a more detailed understanding and significantly improves detection performance.

\begin{figure}[!ht]
\centering
\begin{tcolorbox}[width=0.97\columnwidth, colback=blue!5!white, colframe=blue!75!black, title=A Sample Textual Explanation, left=1.5mm, right=1.5mm, top=0.2mm, bottom=0.2mm, boxsep=0mm]
\{B3,B4\}: The window blinds have uneven spacing, and the light passing through does not align properly with the individual slats, which suggests an error in rendering light and shadows.
\{D1,D2\}: The oven appears to have a distorted handle, and the reflection and shadow around it don't conform to the expected perspective and lighting.
\{D3\}: The drawer underneath the stove has irregular handles that are asymmetrical, which is not typical for kitchen design and could be an oversight by the AI.
\end{tcolorbox}
\vspace{-0.2cm}
\caption{An example for textual explanations, each corresponding to specific patches in the image.}
\label{explanations}
\end{figure}

\begin{figure*}[!htb]
\centering
\resizebox{0.95\textwidth}{!}{
\begin{tabular}{cc}
\includegraphics[width=15cm]{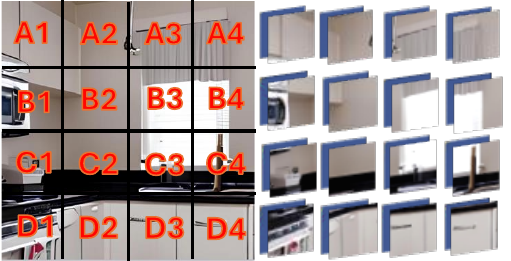}
&
\includegraphics[width=15cm]{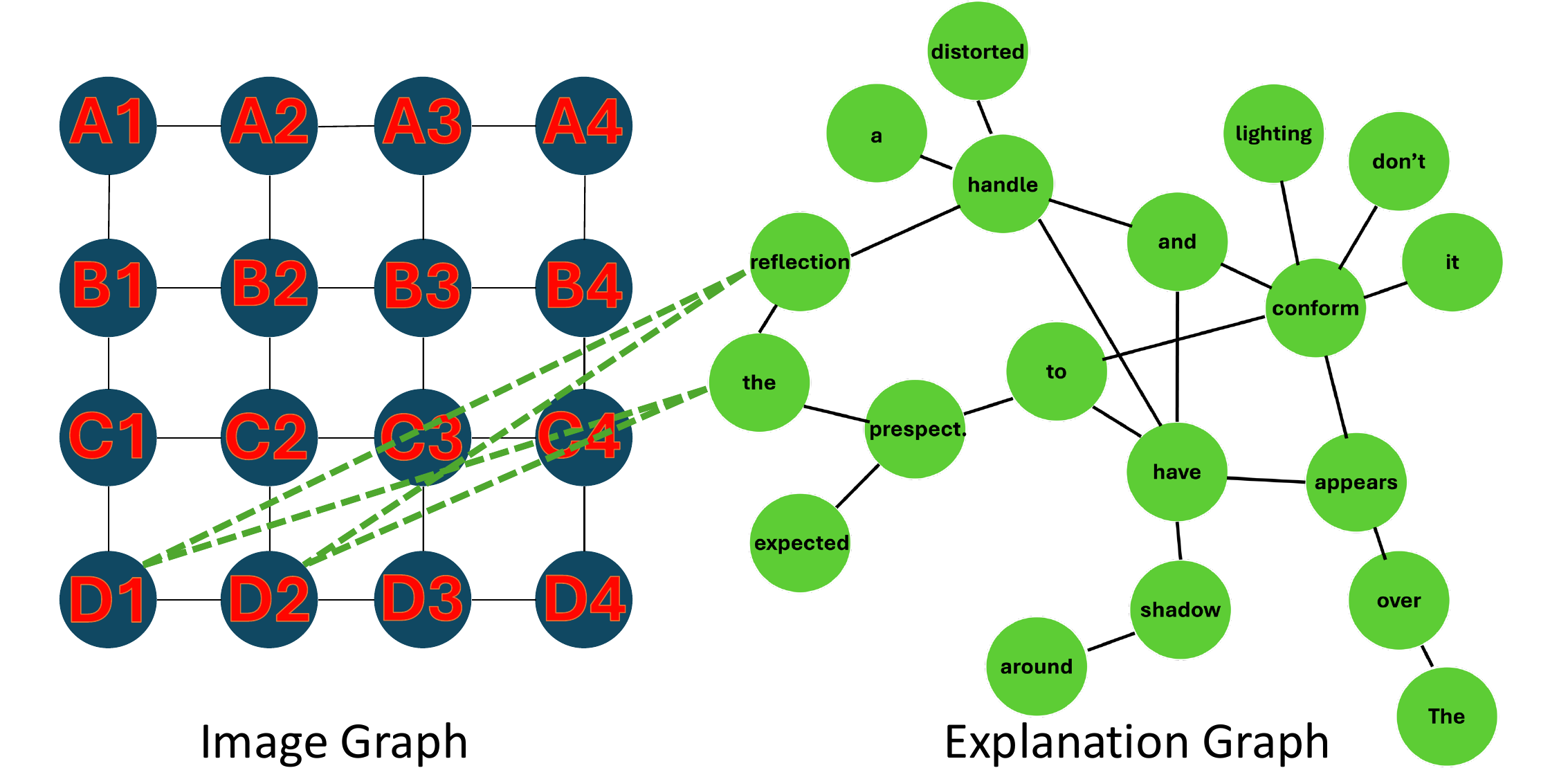}
\\
\Large{(a)}
&
\Large{(b)}\\
\end{tabular}}
\vspace{-0.1cm}
\caption {A sample image graph and its corresponding explanation graph construction and integration: (a) The image with the applied grid (left) alongside the patches and their corresponding DCT (right), and (b) The image graph (left) alongside a sample explanation graph (right), illustrating only connecting 2 nodes (all nodes in the explanation graph are connected to the corresponding patch nodes in the actual implementation).}
\label{img graph}
\end{figure*}
\subsection{Explanation-Patch Integrated Graph Construction}

\par \textbf{Explanation Generation.} The generation of explanations to determine whether an image is real or fake using the VLLM is achieved through a process called \textit{visual prompting}. In \texttt{ViGText}, this involves overlaying the image with a grid of equally sized square patches, each systematically labeled (e.g., A1, A2, A3, A4, B1, etc.). Both the overlaid image and the original image are fed to the VLLM, which enables it to produce explanations that are directly linked to these localized areas. Fig. \ref{grid} illustrates this grid overlay. This segmentation framework ensures accurate integration with the image graph, that links explanations to corresponding patches. Sample generated explanations are shown in Fig. \ref{explanations}, while the prompt template is provided in Fig. \ref{prompt} in Appendix \ref{appendixA}.
\par

Grid-based explanations are essential because, while a single full-image explanation can describe the whole scene, it doesn't clarify how to associate parts of the explanation with specific image regions for graph construction.
The grid ties each explanation to a patch, enabling precise cross-modal edges. However, this introduces a trade-off: smaller patches capture fine details but may lose global context, while larger patches do the opposite. \texttt{ViGText} balances this by choosing intermediate patch sizes and uses the GNN’s message passing to merge local and global cues into coherent, human-understandable reasoning. We analyze this trade-off in more detail in Section\ref{Section5}.

It is important to note that \texttt{ViGText}, in its design philosophy, does not rely on the VLLM as a standalone trust anchor. Rather, it uses the VLLM as a fully local, defender-controlled component, avoiding risks tied to external or opaque models. The VLLM provides fine-grained textual descriptions of lighting, geometry, and texture, which are cross-verified against visual patch features within the dual graph structure. During training, the GNN learns from both matching and deliberately mismatched image-explanation pairs, enabling it to detect inconsistencies across modalities. This ensures that \texttt{ViGText} can reveal manipulations through statistical discrepancies between textual and visual cues, even under adversarial scenarios, making the overall system more robust and reliable.

\par \textbf{Image Graph Construction.} After generating the explanations, \texttt{ViGText} construct a graph that represents patches of the image and its corresponding explanations. The process begins by building the image graph, where each node represents a patch of the image, and nodes corresponding to adjacent patches are connected by undirected edges, as illustrated in the left-hand side of Fig. \ref{img graph}(b). To represent each patch as a node, \texttt{ViGText} utilizes ConvNeXt-Large \cite{liu2022convnet}, a foundation image feature extraction model trained on a subset of the LAION-5B dataset \cite{schuhmann2022laion}. This model extracts feature embeddings for each patch. Additionally, the DCT-transformed patch, illustrated in the right hand side of Fig. \ref{img graph}(a), is passed through the same feature extraction model to produce an embedding. Finally, the two embeddings (image and DCT-based) are averaged to create a robust and comprehensive feature representation for the patch, which is then assigned to the corresponding node in the graph. This dual-domain representation enhances the graph’s ability to capture both spatial and frequency-based artifacts, which are crucial for detecting subtle manipulations in deepfake images.

\par \textbf{Text Graph Construction.} To represent the explanations as graphs, each word in the sentence is depicted as a node, and edges between nodes reflect the grammatical relationships among the words, which are extracted using the dependency parser from spaCy \cite{spacy2}. This structure illustrates how the words interact within the sentence. This method not only captures the roles of the words but also their interactions, resulting in a comprehensive and structured representation of the explanations. \texttt{ViGText} uses Jina \cite{gunther2023jina}, an embedding model, to extract features for the words, assigning each node its corresponding embedding. Finally, \texttt{ViGText} integrates these explanation graphs with the image graph by connecting each node in the explanation graph to the corresponding patch node in the image graph. Fig.~\ref{img graph}(b) shows a sample explanation graph integrated with its patch nodes.

\begin{algorithm}[!htb]
\small
\caption{Patch and Explanation Integrated Graph Construction}
\label{algo}
\begin{algorithmic}[1]
\State \textbf{Input:} An image \texttt{I}, image feature extraction model \texttt{M}, word feature extraction model \texttt{B}, VLLM.
\State \textbf{Output:} Patch and Explanation Word Correspondence Graph
\State Overlay \texttt{I} with the grid mask to produce the image \texttt{Q}. \label{alg:line3}
\State Query the VLLM with \texttt{Q} to produce patch-specific explanations. \label{alg:line4}
\State Split \texttt{I} into patches and extract spatial features for each patch using \texttt{M}. \label{alg:line5}
\State Apply the DCT transformation to each patch and extract frequency-domain features using \texttt{M}. \label{alg:line6}
\State Average the spatial and frequency-domain features to create combined embeddings for each patch. \label{alg:line7}
\State Construct the image graph with nodes representing patches and features corresponding to their combined embeddings. \label{alg:line8}
\State Construct a graph for each explanation with nodes representing words and edges based on grammatical relationships, extracting word features using \texttt{B}. \label{alg:line9}
\State Integrate the image graph with the explanation graphs by connecting each explanation graph node to the corresponding patch node in the image graph. \label{alg:line10}
\State \textbf{Return} the unified graph containing the image graph and the explanation graphs. \label{alg:line11}

\end{algorithmic}
\end{algorithm}

\par The process of constructing the Patch and Explanation Integrated Graph, detailed in Algorithm \ref{algo} and illustrated in Figures \ref{approach} and \ref{pipeline}, begins by overlaying the image \texttt{I} with a grid mask to produce a modified image \texttt{Q} (Step \ref{alg:line3}). The modified image is used to query the VLLM, which generates patch-specific explanations associated with the grid regions (Step \ref{alg:line4}). The image is split into patches, and spatial features are extracted for each patch using the model \texttt{M} (Step \ref{alg:line5}). Additionally, each patch undergoes a DCT, and frequency-domain features are extracted using the same model (Step \ref{alg:line6}). The spatial and frequency-domain features are averaged to form the final feature embeddings for each patch (Step \ref{alg:line7}). These embeddings are used to construct the image graph, with nodes representing patches and edges connecting adjacent patches (Step \ref{alg:line8}). For each explanation, a graph is created where nodes correspond to individual words, and edges encode their grammatical relationships. Word features are extracted using the word embedding model \texttt{B} (Step \ref{alg:line9}). The explanation graphs are then integrated with the image graph by connecting word nodes to their corresponding patch nodes based on the spatial association between explanations and patches (Step \ref{alg:line10}). The unified graph, which combines both the visual and textual data, is returned as the output (Step \ref{alg:line11}).

\section{Experiments}
\label{Section5}

\par In this section, we evaluate the performance of \texttt{ViGText} through a series of  experiments designed to address the questions summarized in Table~\ref{summarytable}. Our results demonstrate that \texttt{ViGText} achieves state-of-the-art detection performance in terms of multiple classification metrics. Moreover, \texttt{ViGText} exhibits strong generalization capabilities, which handle datasets derived from various fine-tuned generative models. Notably, \texttt{ViGText} demonstrates robustness against both foundation model-powered adversarial attacks and targeted attacks crafted with substantial knowledge of its mechanisms. Additionally, the system’s performance remains resilient to variations in design choices, indicating a degree of flexibility in its configuration. Overall, \texttt{ViGText} achieves these advancements with a computational cost that remains comparable to, or tolerable in relation to, existing state-of-the-art approaches. The source code and information about the datasets used in this work can be found in the following repository: \href{https://github.com/AhmadALBarqawi/ViGText}{ViGText}.

\begin{table}[htb]
\centering
\caption{A summary of research questions and key answers.}
\resizebox{0.99\columnwidth}{!}{
\begin{tabular}{|c|l|l|}
\hline
\textbf{Q} & {Property Investigated} & {Key Result} \\ \hline
1 & Detection effectiveness & Highly effective \\ \hline
2 & Generalization & Strong generalization \\ \hline
3 & Robustness & High robustness \\ \hline
4 & Sensitivity to design choices & Generally insensitive \\ \hline
5 & Empirical costs & Tolerable \\ \hline
\end{tabular}
}
\label{summarytable}
\end{table}

\subsection{The Setup, Dataset, and Baselines} 

\par We use the datasets introduced by Sifat et al. \cite{abdullah2024analysis}, which address critical limitations in existing deepfake research. Specifically, \cite{abdullah2024analysis} highlights the lack of control over content and image quality in many existing datasets, which can lead to overestimated performance for state-of-the-art detection methods. To mitigate this issue, \cite{abdullah2024analysis} constructs two carefully curated datasets designed to provide improved control and enable a more accurate evaluation of deepfake detection approaches.
\begin{list}{$\bullet$}{\leftmargin=0em \itemindent=1em}
\item The Stable Diffusion (SD) Dataset contains real images from the LAION-AESTHETICS dataset \cite{laionLAIONAestheticsLAION} and fake images generated using the Realistic Vision v1.4 model \cite{huggingfaceSG161222Realistic_Vision_V14Main}. The dataset spans a broad range of content types, including people, nature, objects, illustrations, and digital art. It is structured to ensure balance, with 16,000 images for training, 2,000 for validation, and 2,000 for testing, equally divided between real and fake images. A key focus of this dataset is to evaluate the generalization of deepfake detection approaches to images generated by fine-tuned variants of generative models, as generalization remains a persistent challenge for existing methods. To assess this, the dataset includes 16 additional testing sets derived from the base SD 1.5 model. Of these, 8 testing sets feature images generated using the Full Model (FM) fine-tuning approach—where all parameters are updated—and the remaining 8 are created using Low-Rank Adaptation (LoRA) fine-tuning \cite{hu2021lora}, a computationally efficient method that updates only a subset of the model’s parameters. Remarkably, even without specific adversarial intent, the images from these fine-tuned variants cause significant performance degradation in many state-of-the-art detection methods, as evidenced by \cite{abdullah2024analysis}. This highlights the growing threat posed by the democratization of fine-tuning techniques and the urgent need for robust, generalizable detection systems. Furthermore, we extend the testing on generalization by creating 8 additional testing sets corresponding to 8 new LoRA fine-tuned variants of the Stable Diffusion 3.5 model \cite{AI_2024}, the current state-of-the-art open-sourced generative model. We choose to include only LoRA fine-tuned variants in this extension due to the significantly larger size of the Stable Diffusion 3.5 model at 8 billion parameters. While LoRA \cite{hu2021lora} fine-tuning is computationally efficient, fine-tuning all the parameters of such a large model would require substantial computational resources that are impractical for most use cases. Table \ref{SDmodels} in Appendix \ref{appendixB} contains more information about the Stable Diffusion 3.5 LoRA models \cite{AI_2024}. This extension provides a comprehensive evaluation of detection methods under challenging scenarios posed by the latest advancements in generative AI, further emphasizing the necessity for detection systems to be both robust and adaptive.

\item StyleCLIP Dataset: This dataset consists of a balanced collection of real and fake face images, specifically designed to study robustness against adversarial attacks using vision foundation models. Real images are sourced from the Flickr-Faces-HQ (FFHQ) dataset \cite{karras2019style}, a high-quality collection of face images, while fake images are generated using StyleGAN2 \cite{karras2020analyzing}, a widely adopted generative model for face synthesis. The dataset is balanced, with 16,000 images for training, and 2,000 images each for validation and testing. Unlike the SD dataset, the StyleCLIP dataset emphasizes robustness evaluation under adversarial scenarios. As detailed in \cite{abdullah2024analysis}, adversarial attacks in this context involve manipulating the semantic properties of face images, such as altering expressions, adding accessories, or modifying facial attributes, without introducing perceptible noise. These attacks use vision foundation models as surrogates to optimize manipulations, enabling the generation of adversarial deepfakes that evade detection. To thoroughly evaluate this challenge, three state-of-the-art foundation models, EfficientNet \cite{tan2019efficientnet}, ViT \cite{dosovitskiy2020image}, and CLIPResNet \cite{radford2021learning}, are employed as surrogate models. These models are used to create three additional adversarial testing sets, each tailored to exploit the weaknesses of existing deepfake detection approaches. This dataset, therefore, provides a critical benchmark for assessing the resilience of detection systems in adversarial settings, highlighting the vulnerabilities exposed by foundation model-powered attacks. 
\par To extend our evaluation, we implemented a more advanced adversarial attack that simulates an attacker with significant knowledge of \texttt{ViGText}. This hypothetical attacker is assumed to possess detailed insights into the training dataset and the graph creation pipeline used by \texttt{ViGText}. Based on these assumptions, the attacker creates a surrogate model designed to mimic \texttt{ViGText}’s functionality. The surrogate model comprises two Graph Convolutional Layers and uses the Dinov2 \cite{oquab2023dinov2} and Jina \cite{gunther2023jina} foundation models for extracting image and word embeddings, respectively. This surrogate is trained on the StyleCLIP dataset, achieving high performance across classification metrics, with accuracy, recall, precision, and F1 scores all exceeding 95\%. Using this surrogate model, we further train the StyleGAN2 \cite{karras2020analyzing} generative model to produce adversarial images. These images are specifically crafted to evade detection by the surrogate and, consequently, by \texttt{ViGText}. The attack optimizes the generator by minimizing the cross-entropy loss between the surrogate logits z and the target label y, with y chosen as the label for real images so that the generator produces evasive adversarial examples:
\begin{equation} 
\mathcal{L}_{\text{adv}} = - \mathbb{E}_{\mathbf{x} \sim G(\theta)} \left[ y \log p(\mathbf{z}) + (1 - y) \log (1 - p(\mathbf{z})) \right],
\end{equation}
where $G(\theta)$ is the StyleGAN2 generator with parameters $\theta$ , p(z)  represents the surrogate model’s output probabilities for the real label, and  x $\sim G(\theta)$  are the generated images. This setup allows us to evaluate \texttt{ViGText} against adversarial images crafted by an attacker that closely mimics the real-world threat of a knowledgeable adversary.
\end{list}

\par It is important to note that while the StyleCLIP dataset and its surrogate-based attacks effectively test robustness under strong visual manipulations, they do not account for coordinated attacks that simultaneously target both the image and the VLLM-generated explanations. Executing such dual-objective attacks would require white-box access to both the image generator and the VLLM, enabling gradient-based optimization across components. This substantially increases the complexity and computational cost, placing it beyond the scope of our current evaluation and leaving it as an open direction for future research.

\par We report classical classification metrics for all experiments conducted in this section, which include accuracy, precision, recall, and f1 scores. As for the baselines, we select 3 of the state-of-the-art approaches that perform decently in the analysis in \cite{abdullah2024analysis}. These approaches are:
\begin{list}{$\bullet$}{\leftmargin=0em \itemindent=1em}
\item DCT \cite{ricker2022towards}: This approach works by extracting frequency-domain features from images using a discrete cosine transform (DCT) to identify subtle artifacts. These features are log-scaled for better performance and then used to train a Logistic Regression classifier, which effectively differentiates between real and fake images.

\item DE-FAKE \cite{sha2023fake}: This approach builds a deepfake detector using the CLIP model \cite{radford2021learning} by augmenting the image’s embedding with the embedding of the text prompt used to generate the image. These augmented embeddings are used to train a 2-layer multilayer perceptron as a classifier.

\item UnivCLIP \cite{ojha2023towards}: This recent approach utilizes a large foundation model, specifically the CLIP:ViT-L/14 model \cite{radford2021learning}. This approach extracts features from the frozen CLIP:ViT model and then uses either a nearest neighbor classifier or a linear classification layer, with further training, to determine if an image is real or fake. The linear classifier is preferred here for better performance.
\end{list}

While recent approaches such as ObjectFormer \cite{wang2022objectformer} and detectors built on large-scale vision-language models (VLMs) \cite{khan2024clipping,chang2023antifakeprompt} demonstrate promising detection capabilities, they also depend on resource-intensive architectures like dense transformer attention or billion-parameter language models. These designs face scalability limitations that hinder their practicality for widespread deployment. In contrast, our focus is on methods that balance strong detection performance with computational efficiency. \texttt{ViGText} demonstrates state-of-the-art accuracy and robustness on up-to-date, challenging datasets reflecting modern generative techniques, while maintaining a lightweight graph-based architecture, which ensures broader applicability and scalability.

\par For \texttt{ViGText}, we use a consistent GNN architecture across all experiments. The model comprises three Graph Attention Network (GAT) \cite{velivckovic2017graph} layers with two attention heads, followed by batch normalization and ReLU activation after each layer. Dropout is applied after each layer to prevent overfitting by regularizing the training process. The node features are aggregated using global mean pooling before being passed to a fully connected layer for final classification.

\par The model is trained for 40 epochs using the Adam optimizer \cite{kingma2014adam} to minimize Cross Entropy loss, with learning rate scheduling employed to adjust the learning rate dynamically during training. All experiments are conducted with a 4x4 patch size and utilize Qwen2-VL-7B-Instruct \cite{wang2024qwen2} as the explanation-generating VLLM. The experiments are performed on a workstation equipped with 64 GB of RAM, an 8 GB RTX 2070 GPU, and a 32-core Intel Xeon processor.

\subsection{Performance Analysis}
\begin{table}[!thb]
 \centering
 \caption{Performance analysis on the respective testing sets of the datasets (The highest is in bold).}
 \resizebox{0.95\columnwidth}{!}{%
 \begin{tabular}{|l|c|c|c|c|c|c|c|c|}
 \hline
 \multirow{2}{*}{Approach} & \multicolumn{4}{c|}{SD} & \multicolumn{4}{c|}{StyleCLIP} \\ \cline{2-9}
 & Accuracy & Precision & Recall & F1 & Accuracy & Precision & Recall & F1 \\ \hline
 DCT & 85.50 & 83.30 & 88.80 & 85.96 & 98.80 & 98.22 & \textbf{99.40} & 98.80 \\ \hline
 DE-FAKE & 92.45 & 91.17 & 94.00 & 92.5 & 74.05 & 75.34 & 71.50 & 73.37 \\ \hline
 UnivCLIP & 93.04 & 92.33 & 93.89 & 93.10 & 93.04 & 93.79 & 92.19 & 92.99 \\ \hline
 \texttt{ViGText} & \textbf{99.25} & \textbf{99.8} & \textbf{98.52} & \textbf{99.26} & \textbf{99.60} & \textbf{99.90} & 99.21 & \textbf{99.60} \\ \hline
 \end{tabular}%
 }
 \label{tab:defense_comparison}
\end{table}

\par We begin by addressing \textbf{Q1: How effective is \texttt{ViGText} at detecting deepfakes?} To evaluate this, we compare the performance of \texttt{ViGText} against the baselines mentioned above using the specified quality metrics across both datasets. As shown in Table \ref{tab:defense_comparison}, \texttt{ViGText} consistently outperforms the latest state-of-the-art techniques, demonstrating a strong capability to detect deepfakes generated by various approaches and reflecting the practical threat landscape of deepfake techniques.

\par These results highlight the effectiveness of \texttt{ViGText}’s unique integration of visual and textual information through a dual-graph structure. Through the use of spatial and frequency embeddings, as well as detailed context-aware textual explanations, \texttt{ViGText} achieves superior detection performance. Additional experiments on images from state-of-the-art diffusion APIs and under advanced adversarial attacks are reported in Appendix~\ref{appendixC}, and further sample cases only detected by \texttt{ViGText} are illustrated in Appendix~\ref{appendixD}.

\begin{figure}[!htb]
\centering
\resizebox{0.99\columnwidth}{!}{
\begin{tabular}{c}
\includegraphics[width=20cm]{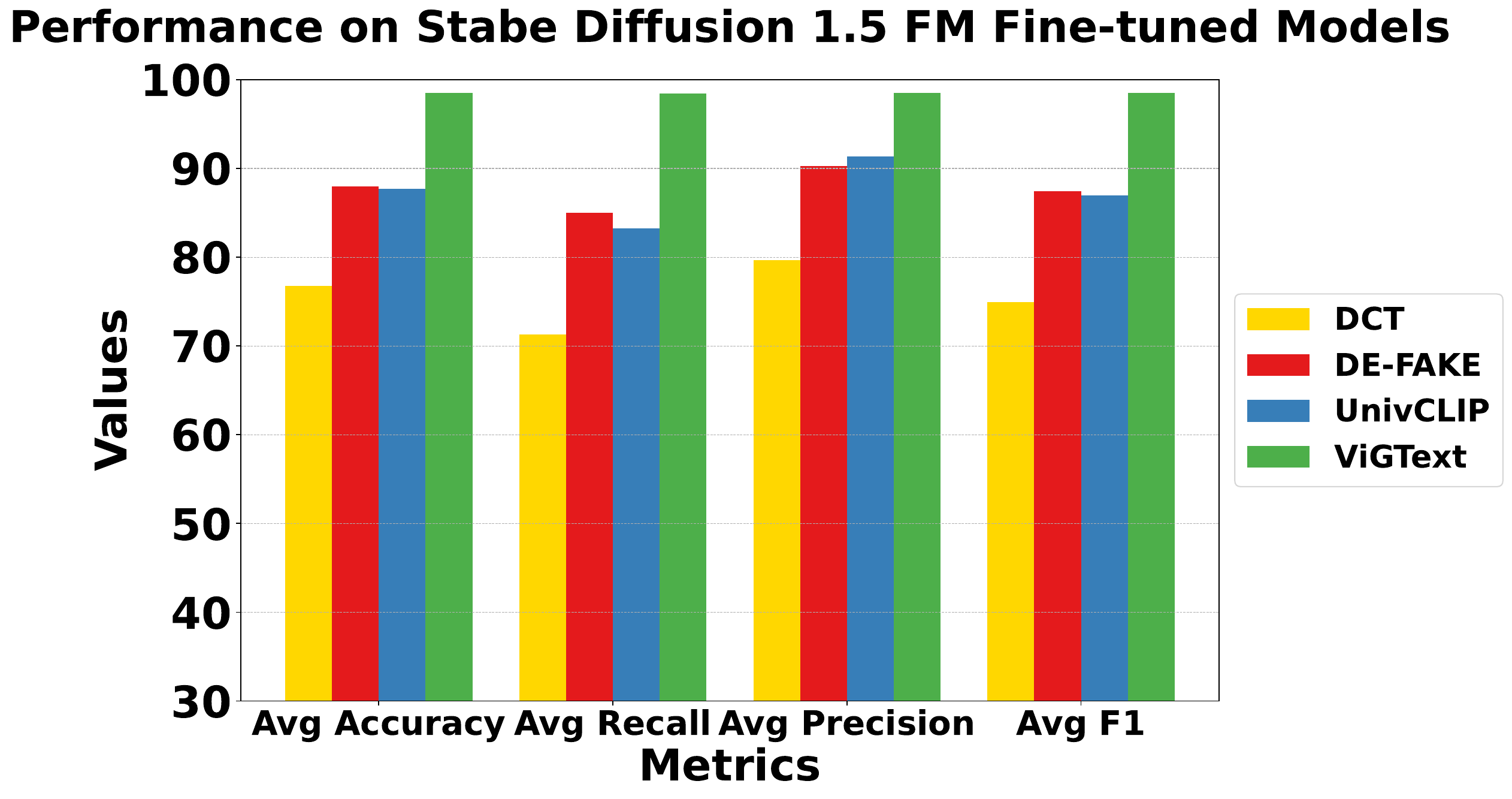}
\\
\Huge{(a)}
\\
\includegraphics[width=20cm]{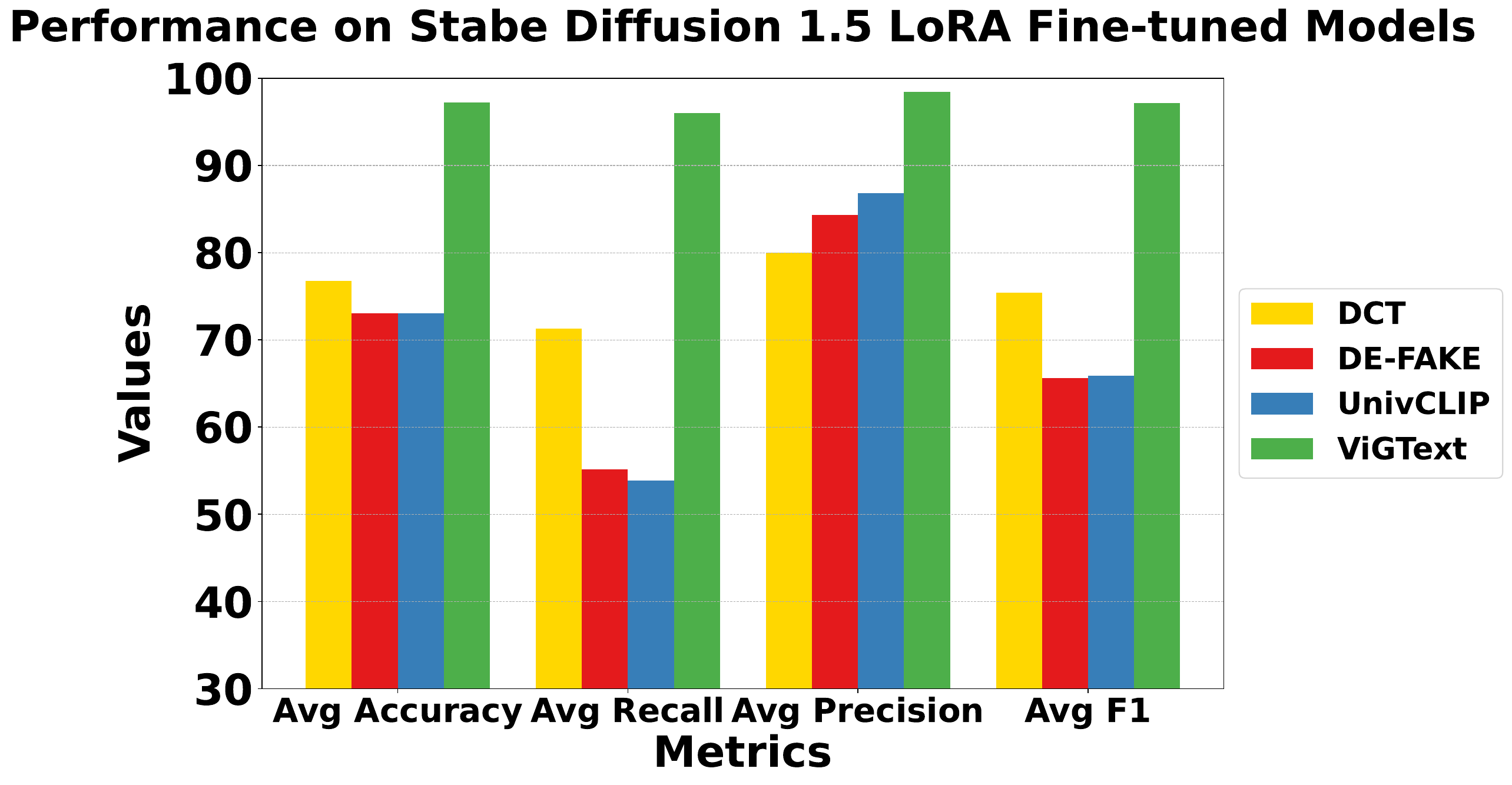}
\\
\Huge{(b)}
\\
\includegraphics[width=20cm]{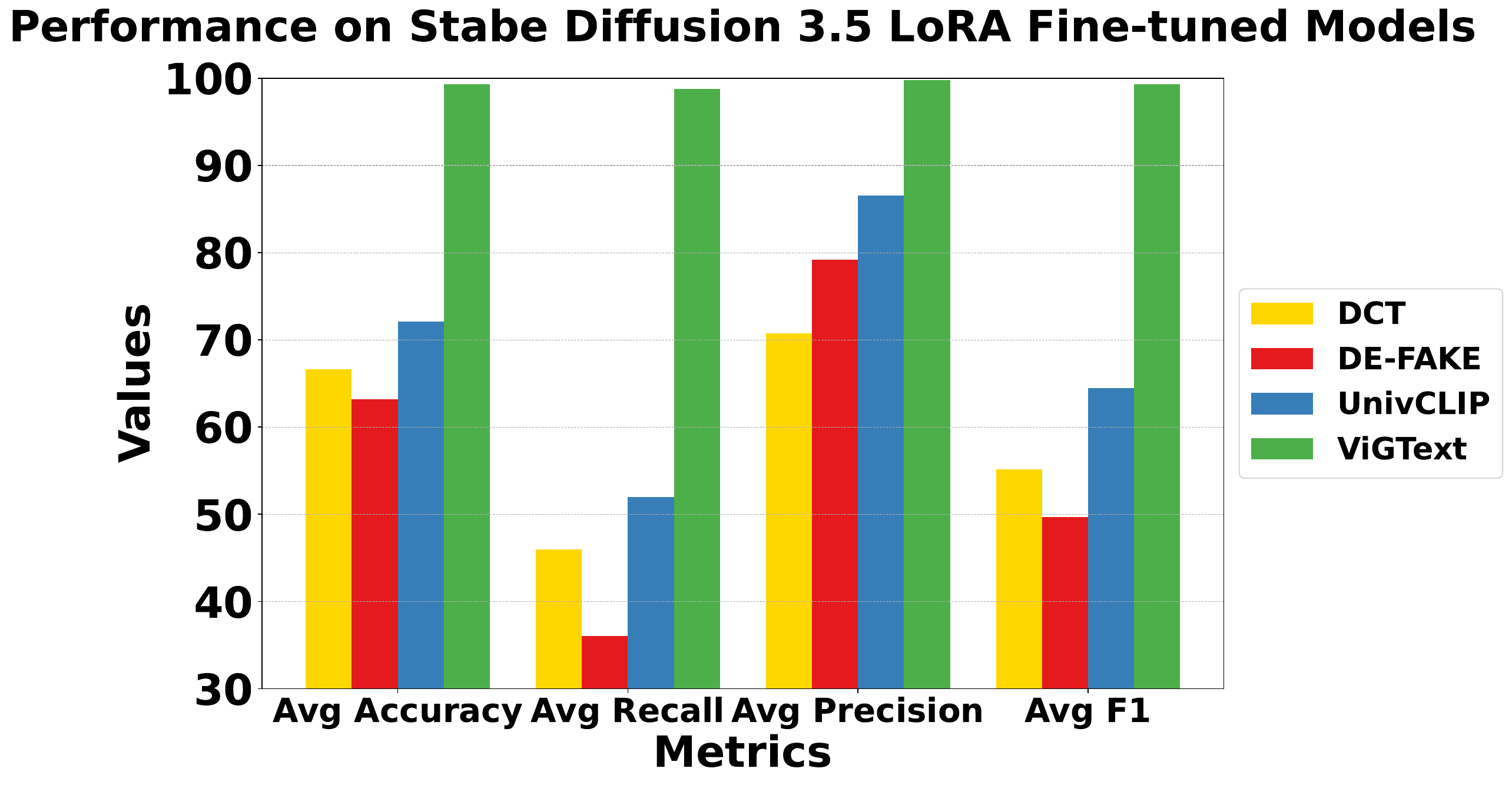}
\\
\Huge{(c)}
\\
\end{tabular}}
\vspace{-0.2cm}
\caption{Generalizing performance on (a) Stable Diffusion 1.5 FM fine-tuned models, (b) Stable Diffusion 1.5 LoRA fine-tuned models and (c) Stable Diffusion 3.5 LoRA fine-tuned models.}
\label{Gen} 
\end{figure}

\subsection{Generalization}

\par Here, we address \textbf{Q2: Can \texttt{ViGText} generalize well enough to detect images generated by various fine-tuned variants?} This evaluation focuses primarily on the SD dataset. \texttt{ViGText} is trained using the training data from this dataset and tested on 24 separate testing sets corresponding to fine-tuned variants of different SD models. These testing sets are split between Full Model (FM) fine-tuned and LoRA fine-tuned variants. Fig. \ref{Gen}(a) shows the average performance metrics across the 8 FM fine-tuned variants of the SD 1.5 model, while Fig. \ref{Gen}(b) and Fig. \ref{Gen}(c) illustrate the results for the 16 LoRA fine-tuned variants of the SD 1.5 and 3.5, respectively.

\begin{figure*}[!thb]
\centering
\resizebox{0.92\textwidth}{!}{
\includegraphics[width=12cm]{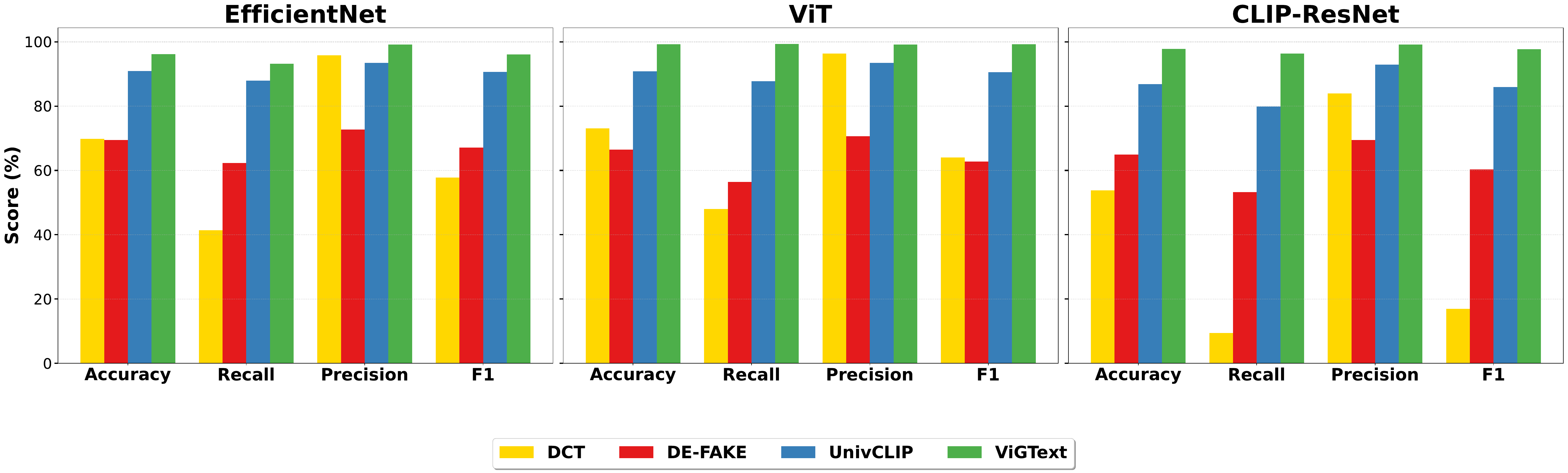}
}
\vspace{-0.3cm}
\caption {Bar plots illustrating the robustness performance of \texttt{ViGText} and baseline approaches on adversarially manipulated testing sets of the StyleCLIP dataset. Each plot corresponds to a different surrogate foundation model—EfficientNet, ViT, and CLIP-ResNet—used to craft the adversarial manipulations.}
\label{robust}
\end{figure*}
\par As illustrated in Fig. \ref{Gen}, \texttt{ViGText} demonstrates superior generalization performance compared to baseline methods across all metrics, which showcases its ability to detect fake images generated by diverse fine-tuned models. This is particularly significant given that traditional data-driven approaches often exhibit decreased performance when tested on data with altered distributions, such as those arising from fine-tuned generative models. In contrast, \texttt{ViGText} mitigates this limitation through its graph-based framework, which focuses on learning the structural topologies of the graphs derived from data points. Additionally, \texttt{ViGText} uses frequency domain features that are content-agnostic, capturing subtle features that are invariant to the content of the images. These features play a critical role to enhance generalization as they reduce dependence on the underlying data distribution or the characteristics of the generative models used during training. \texttt{ViGText} combines its graph-based architecture with frequency domain analysis, to effectively adapt to varied and challenging test scenarios, which reinforces generalization.

\subsection{Robustness}

\par In this subsection, we address \textbf{Q3: How robust is \texttt{ViGText} against foundation model-based adversarial attacks and manipulations on images?} For this purpose, we focus on the StyleCLIP dataset and its associated adversarially manipulated testing sets. These testing sets are crafted using a state-of-the-art adversarial attack \cite{abdullah2024analysis}, which uses foundation models trained on extensive datasets to generate manipulations that reduce the performance of conventional detection methods. The manipulations aim to exploit subtle semantic properties of the images, making the attacks more effective without introducing perceptible noise.

\par We train \texttt{ViGText} on the training portion of the StyleCLIP dataset, then test on these manipulated sets, with each testing set corresponding to a different surrogate foundation model, EfficientNet, ViT, and CLIP-ResNet, used to generate the adversarial attacks. As shown in Fig. \ref{robust}, \texttt{ViGText} consistently outperforms other methods across all evaluated metrics. This demonstrates that \texttt{ViGText} exhibits significantly less performance degradation when faced with adversarial attacks compared to baseline approaches. The resilience of \texttt{ViGText} is particularly noteworthy given the sophistication of the attacks, which utilize foundation models trained on millions of diverse images to approximate high-quality surrogates. This inherent robustness makes \texttt{ViGText} an effective solution for combating adversarial manipulations.

\par Next, we evaluate the robustness of \texttt{ViGText} in a scenario where the adversary has substantial knowledge about the detection system. The adversary is assumed to have access to the same training dataset (StyleCLIP dataset) and the same pipeline used for to construct graph structures between images and textual explanations. This setup models a highly capable and informed adversary to rigorously assess \texttt{ViGText}’s resilience under such a challenging threat model. To simulate this scenario, we design a surrogate detection model with a reasonably chosen architecture that mimics the characteristics of \texttt{ViGText}. The surrogate model is trained using the StyleCLIP dataset and achieves over 95\% on all detection metrics (accuracy, precision, recall, and F1 score) when evaluated on the StyleCLIP testing set. This high-performing surrogate is then used to generate adversarial images by optimizing the StyleGAN2 \cite{karras2020analyzing} generator to minimize the cross-entropy loss between the surrogate’s logits and the target label (real images), which effectively crafts evasive examples.

\par We test \texttt{ViGText} on these adversarial images, and it achieves an accuracy of \textbf{95.85\%}, recall of \textbf{91.7\%}, precision of \textbf{99.2\%}, and an F1 score of \textbf{95.67\%}, compared to the original metrics of accuracy \textbf{99.6\%}, precision \textbf{99.9\%}, recall \textbf{99.21\%}, and F1 score \textbf{99.6\%}. While these adversarial images have a greater impact on \texttt{ViGText}’s performance compared to foundation model-based attacks, this outcome is expected as the surrogate closely mirrors the actual detection system, which provides the adversary with a considerable advantage.

\par The previous results highlight \texttt{ViGText}’s robustness, even against an adversary with significant knowledge of its design and training data. The graph-based framework, integration of frequency-domain features, and effective combination of visual and textual information enable \texttt{ViGText} to maintain strong detection performance under this extreme threat model. This demonstrates \texttt{ViGText}'s capability to withstand attacks from resourceful and well-informed adversaries.

\begin{table}[!htb]
 \centering
 \Large
 \caption{Performance with SD as the test set across different image resolutions (The highest is in bold).}
 \resizebox{1\columnwidth}{!}{ 
 \begin{tabular}{|c|c|c|c|c|c|c|c|c|c|c|c|c|}
 \hline
 \multirow{2}{*}{Resolution} & \multicolumn{4}{c|}{450x450} & \multicolumn{4}{c|}{512x512} & \multicolumn{4}{c|}{550x550} \\ \cline{2-13}
 & Acc & Prec & Rec & F1 & Acc & Prec & Rec & F1 & Acc & Prec & Rec & F1 \\ \hline
 DCT & 46.60	&6.20	&32.29&	10.40 & 85.50 & 83.30 & 88.80 & 85.90 & 51.60&	6.20	&67.39	&11.35 \\ \hline
 DE-FAKE & 93.40	&93.90	&92.97&	93.43 & 92.40 & 91.10 & 94.00 & 92.50 & 93.45	&94.10	&92.89	&93.49 \\ \hline
 UnivCLIP & 92.29&	92.09	&92.46	&92.28 & 93.00 & 92.30 & 93.90 & 93.10 & 92.44	&92.19&	92.66	& 92.43 \\ \hline
 \texttt{ViGText} & \textbf{96.40} & \textbf{99.90}	& \textbf{93.28}	& \textbf{96.53}& \textbf{99.25} & \textbf{99.80} & \textbf{98.52} & \textbf{99.26} & \textbf{97.20}	& \textbf{95.20}	& \textbf{99.17} & \textbf{97.14} \\ \hline
 \end{tabular}
 }
 \label{res}
\end{table}

\par Continuing with the robustness evaluation, we test \texttt{ViGText} against manipulations caused by changes in image resolution. This is an important aspect of robustness, as practical applications often involve input images that vary in resolution due to diverse capture conditions or post-processing steps. Table \ref{res} presents the results on the SD dataset, which originally consists of 512x512 resolution images. While Table \ref{resStyle} shows the corresponding results for the StyleCLIP dataset, which originally consists of 1024x1024 resolution images.

\par The results in Table \ref{res} demonstrate that \texttt{ViGText} achieves minimal degradation in performance across all tested resolutions on the SD dataset. This consistent performance highlights its robustness to resolution changes, with accuracy, precision, recall, and F1 scores remaining high. Notably, \texttt{ViGText} outperforms all baseline methods at each resolution, which suggests that its graph-based framework contributes to its ability to adapt effectively to variations in image resolution.

\begin{table}[!htb]
 \centering
 \Large
 \caption{Performance with SD as the test set across different geometric warp operations (The highest is in bold).}
 \resizebox{\columnwidth}{!}{ 
 \begin{tabular}{|c|c|c|c|c|c|c|c|c|c|c|c|c|}
 \hline
 \multirow{2}{*}{Technique} & \multicolumn{4}{c|}{Rotate} & \multicolumn{4}{c|}{Scale and Translate} \\ \cline{2-9}
 & Acc & Prec & Rec & F1 & Acc & Prec & Rec & F1 \\ \hline
 DCT & 54.6 & 54.9 & 51.4 & 53.0 & 57.5 & 62.5 & 37.4 & 46.8 \\ \hline
 DE-FAKE & 86.8 & 81.7 & 95.0 & 87.8 & 89.2 & 86.0 & 93.6 & 89.6 \\ \hline
 UnivCLIP & 88.1 & 84.0 & 94.1 & 88.8 & 90.9 & 86.6 & 96.9 & 91.5 \\ \hline
 \texttt{ViGText} & \textbf{98.0} & \textbf{96.1} & \textbf{100.0} & \textbf{98.0} & \textbf{99.6} & \textbf{99.9} & \textbf{99.2} & \textbf{99.6} \\ \hline
 \end{tabular}
 }
 \label{geowarpSD}
\end{table}

\par Next, Tables \ref{geowarpSD} and \ref{appearwarpSD} present the performance of all evaluated methods on the SD dataset under geometric and appearance-based warp operations, respectively. \texttt{ViGText} demonstrates consistently superior metrics across these transformations, indicating robust resilience to spatial distortions. In particular, \texttt{ViGText} maintains high accuracy, precision, recall, and F1 scores even when subjected to significant geometric modifications, showcasing its adaptability in real-world scenarios where images may be rotated or spatially transformed. For completeness, we also include supplementary experiments that cover variations in resolution as well as geometric and appearance-based warp operations on the StyleCLIP dataset in Appendix~\ref{appendixE}, further demonstrating the consistency of \texttt{ViGText}’s performance across different datasets and manipulation types.

\begin{table}[!htb]
 \centering
 \Large
 \caption{Performance with SD as the test set across different appearance-based warp operations (The highest is in bold).}
 \resizebox{\columnwidth}{!}{ 
 \begin{tabular}{|c|c|c|c|c|c|c|c|c|c|c|c|c|}
 \hline
 \multirow{2}{*}{Technique} & \multicolumn{4}{c|}{Blurring} & \multicolumn{4}{c|}{Brightness} \\ \cline{2-9}
 & Acc & Prec & Rec & F1 & Acc & Prec & Rec & F1 \\ \hline
 DCT & 80.00 & 67.00 & 90.54 & 77.01 & 81.2 & 78.7 & 85.6 & 81.9 \\ \hline
 DE-FAKE & 92.60 & 95.70 & 90.11 & 92.82 & 92.2 & 92.7 & 91.7 & 92.2 \\ \hline
 UnivCLIP & 92.74 & 96.39 & 89.84 & 93.01 & 91.4 & 90.9 & 92.2 & 91.5 \\ \hline
 \texttt{ViGText} & \textbf{97.60} & \textbf{99.90} & \textbf{95.42} & \textbf{97.66} & \textbf{99.6} & \textbf{99.2} & \textbf{99.9} & \textbf{99.6} \\ \hline
 \end{tabular}
 }
 \label{appearwarpSD}
\end{table}

\par Finally, Table \ref{fgsmpdg} reports results for adversarial robustness using the Fast Gradient Sign Method (FGSM) \cite{goodfellow2014explaining} and Projected Gradient Descent (PGD) \cite{madry2017towards} attacks at varying noise levels. Here, \texttt{ViGText} maintains a clear advantage over the other methods, achieving the highest accuracy in all tested conditions. Even at higher noise magnitudes, \texttt{ViGText} exhibits a notable margin of improvement compared to competing approaches. This strong adversarial resilience, combined with its robustness to geometric and appearance-based transformations, confirms the suitability of \texttt{ViGText} for handling diverse and challenging manipulations in practice.

\begin{table}[]
\centering
\Large
\caption{Accuracy against adversarial images using FGSM and PGD attacks (The highest is in bold).}
\resizebox{\columnwidth}{!}{ 
\begin{tabular}{|cc|c|c|c|c|}
\hline
\multicolumn{2}{|c|}{---}                               & \multirow{2}{*}{DCT} & \multirow{2}{*}{DE-FAKE} & \multirow{2}{*}{UnivCLIP} & \multirow{2}{*}{\texttt{ViGText}} \\ \cline{1-2}
\multicolumn{1}{|c|}{Attack}                & Noise ($\epsilon$) &                      &                          &                           &                          \\ \hline
\multicolumn{1}{|c|}{\multirow{3}{*}{FGSM}} & 0.0001    & 82.41                & 88.21                    & 82.37                     & \textbf{96.43}                    \\ \cline{2-6} 
\multicolumn{1}{|c|}{}                      & 0.001     & 75.78                & 83.02                    & 54.99                     & \textbf{93.71}                   \\ \cline{2-6} 
\multicolumn{1}{|c|}{}                      & 0.01      & 71.09                & 71.48                    & 37.19                     & \textbf{89.19}                    \\ \hline
\multicolumn{1}{|c|}{\multirow{3}{*}{PGD}}  & 0.0001    & 35.16                & 63.52                    & 61.84                     & \textbf{91.46}                    \\ \cline{2-6} 
\multicolumn{1}{|c|}{}                      & 0.001     & 16.24                & 61.04                    & 56.55                     & \textbf{87.83}                    \\ \cline{2-6} 
\multicolumn{1}{|c|}{}                      & 0.01      & 9.47                 & 58.36                    & 51.31                     & \textbf{80.94}                    \\ \hline
\multicolumn{1}{|c|}{No Attack}             & ---       & 85.50                & 92.45                    & 93.04                     & \textbf{99.25}                    \\ \hline
\end{tabular}}
\label{fgsmpdg}
\end{table}

\subsection{Sensitivity to Design Choices}
\par In this subsection, we investigate \textbf{Q4: How sensitive is \texttt{ViGText} to the design choices made during its development?} We explore the sensitivity of \texttt{ViGText} to the number of patches into which the image is divided, and evaluate the effect of the number of patches on performance across different datasets and scenarios.
\begin{table}[!htb]
 \centering
 \small 
 \caption{performance on SD dataset and StyleCLIP dataset across different patch sizes}
 \setlength{\tabcolsep}{3pt} 
 \resizebox{\columnwidth}{!}{ 
 \begin{tabular}{|l|c|c|c|c|c|c|c|c|}
 \hline
 \multirow{2}{*}{{Patches}} & \multicolumn{4}{c|}{{SD Dataset}} & \multicolumn{4}{c|}{{StyleCLIP Dataset}} \\ \cline{2-9}
 & {Acc} & {Prec} & {Rec} & {F1} & {Acc} & {Prec} & {Rec} & {F1} \\ \hline
 3x3 & 98.00 & 96.15 & 99.90 & 98.04 & 99.60 & 99.21 & 99.90 & 99.60 \\ \hline
 4x4 & 99.25	&99.80	&98.52	&99.26 & 99.60	&99.90&	99.21	&99.60 \\ \hline
 5x5 & 98.40 & 96.90 & 99.90 & 98.43 & 99.90 & 99.90 & 99.80 & 99.90 \\ \hline
 \end{tabular}
 }
 \label{patchTestSet}
\end{table}
\par Table \ref{patchTestSet} summarizes the results on the SD and StyleCLIP datasets when we vary the patch size. The results show that the performance on both datasets remains relatively stable across patch sizes, indicating that \texttt{ViGText} is not overly sensitive to this parameter. However, moving to the generalization performance on the fine-tuned model datasets in Table \ref{Patchgen}, we observe a notable increase in performance as the patch size decreases (and consequently, the number of patches increases).
\begin{table}[!htb]
 \centering
 \Large
 \caption{Performance on LoRA and FM finetuned models across different numbers of image patches.}
 \resizebox{1\columnwidth}{!}{ 
 \begin{tabular}{|c|c|c|c|c|c|c|c|c|c|c|c|c|}
 \hline
 \multirow{2}{*}{Patches} & \multicolumn{4}{c|}{SD 1.5 LoRA} & \multicolumn{4}{c|}{SD 1.5 FM} & \multicolumn{4}{c|}{SD 3.5 LoRA} \\ \cline{2-13}
 & Acc & Prec & Rec & F1 & Acc & Prec & Rec & F1 & Acc & Prec & Rec & F1 \\ \hline
 3x3 & 92.68	& 98.92 & 91.85 & 95.19 & 97.06 & 98.98 & 97.32 & 98.13 & 98.96 & 99.00 & 99.70 & 99.35 \\ \hline
 4x4 & 97.25 & 96.00 & 98.45 & 97.18 & 98.50 & 98.49 & 98.45 & 98.49 & 99.30	& 99.79 & 98.80 & 99.28 \\ \hline
 5x5 & 98.70 & 99.19 & 99.17 & 99.18 & 99.26 & 99.20 & 99.87 & 99.53 & 99.35 & 99.20 & 99.99 & 99.60 \\ \hline
 \end{tabular}
 }
 \label{Patchgen}
\end{table}
\par
\begin{figure}[!htb]
\centering
\resizebox{0.99\columnwidth}{!}{
\includegraphics[width=12cm]{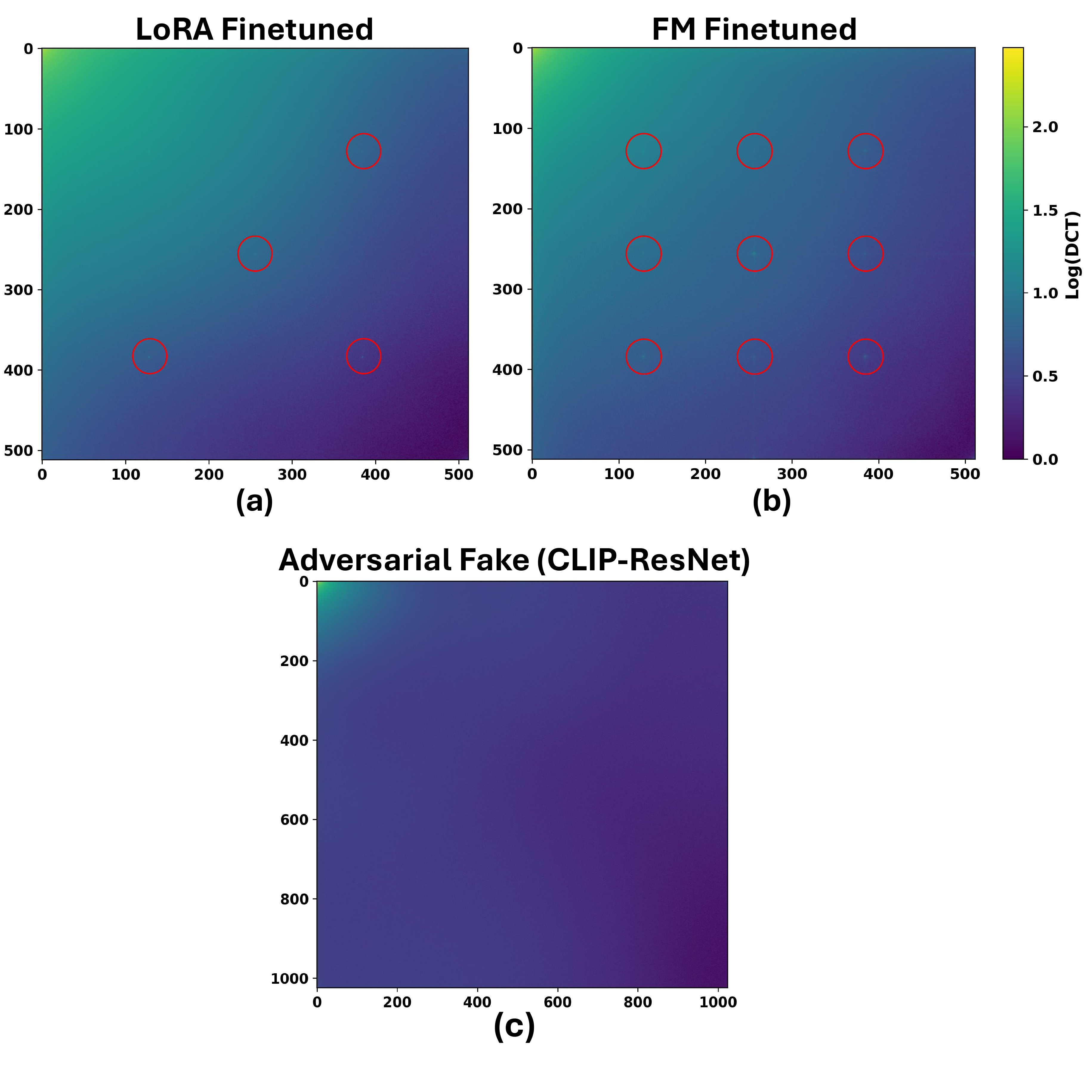}
}
\vspace{-0.9cm}
\caption {Log DCT frequency spectrum showing (a) artifacts for images generated using LoRA fine-tuned models, (b) artifacts for ones generated using FM fine-tuned models and (c) no artifacts in the adversarial generated images. Red circles showcasing the artifacts.}

\label{frequncy}
\end{figure}

 The increase in performance as the number of patches increases can be attributed to the nature of the artifacts present in these images. As shown in Fig. \ref{frequncy}, images generated by LoRA fine-tuned models (Fig. \ref{frequncy}(a)) and FM fine-tuned models (Fig. \ref{frequncy}(b)) contain localized artifacts that are better captured with smaller patches. When we decrease the patch size the spatial granularity of the graph increases, which allows the model to better localize and represent these distortions. Smaller patches also create more graph nodes, which enhances the model’s sensitivity to subtle variations in localized regions, further improving its ability to generalize.
\begin{table}[!htb]
 \centering
 \caption{Average Performance on the 3 adversrially manipulated testing sets of the StyleCLIP dataset across different number of image patches.}
 \begin{tabular}{|c|c|c|c|c|}
 \hline
 Patches & Acc & Prec & Rec & F1  \\ \hline
 3x3 & 99.30 & 99.20 & 99.40 & 99.30  \\ \hline
 4x4 & 97.76 & 96.33 & 99.18 & 97.71  \\ \hline
 5x5 & 86.56 & 98.56 & 74.13 & 83.23  \\ \hline
 \end{tabular}
 \label{Patchrobust}
\end{table}
\par We notice a different behavior, as seen in Table \ref{Patchrobust}, where the performance on adversarially manipulated testing sets of the StyleCLIP dataset exhibits the opposite trend, with larger patch sizes yielding better results. This behavior is likely because adversarial images (Fig. \ref{frequncy}(c)) lack clear artifacts, and smaller patches may fail to capture meaningful features. For these images, the signal within individual patches is weaker, and the increase in the number of graph nodes can dilute the signal, which makes it harder for \texttt{ViGText} to distinguish between real and fake content. Additionally, smaller patches reduce the ability of the graph to capture global patterns, which are crucial to detect adversarial manipulations designed for evasion.

\par The sensitivity of \texttt{ViGText} to patch size is closely tied to the nature of the images being analyzed. For artifact-rich images, smaller patches enhance performance by capturing localized distortions, while for artifact-free adversarial images, larger patches are more effective as they preserve global context. This suggests that an adaptive patching strategy, which dynamically adjusts patch size based on the characteristics of the input image, could further enhance \texttt{ViGText}’s robustness and generalization capabilities.

\subsection{Cost Analysis}

\par Finally, we address \textbf{Q5: How costly is it to run \texttt{ViGText}?} We compare \texttt{ViGText} to UnivCLIP, which demonstrated the second-best overall performance across the previous experiments. The time cost is measured as the total time required to preprocess the image, run the model inference, and, for \texttt{ViGText}, includes graph construction. These results are averaged over 2000 test samples.

\par On average, UnivCLIP takes 1.650 seconds per image from preprocessing to inference, while \texttt{ViGText} takes only slightly more at 1.755 seconds. This marginal increase in time cost, just 0.105 seconds, highlights the efficiency of \texttt{ViGText}. The results demonstrate that \texttt{ViGText} achieves its superior detection performance with minimal additional computational cost. The use of advances in VLLMs, such as Qwen2 \cite{wang2024qwen2}, significantly contributes to this efficiency. Qwen2-VL-7B-Instruct is a compact yet powerful model that provides a deep understanding of image content and generates detailed textual explanations with low computational overhead. Through these advances, \texttt{ViGText} is able to integrate high-quality explanations into its graph-based framework while maintaining a competitive runtime.

\par These findings showcase that \texttt{ViGText} not only excels in detection accuracy and robustness but also remains practical for real-world deployment, as it achieves state-of-the-art performance with a near-negligible increase in time cost.

\section{Related Work}
\label{Section6}
\par \textbf{Deepfake Detection.} Recent deepfake detection approaches can be broadly categorized based on the model architecture into two main approaches: CNN-based approaches and traditional machine learning models-based approaches. CNN-based approaches focus on utilizing deep learning architectures to detect fake images. For instance, \cite{chai2020makes} explores the characteristics of fake images that enable detection across different generative models and datasets. This study introduces a patch-based classifier with limited receptive fields, which emphasizes local image artifacts rather than global structures. \cite{liu2020global} presents Gram-Net, a model that enhances fake face detection by concentrating on global texture features, which improves robustness and generalization across various GAN models and image distortions. \cite{he2021beyond} also aims to improve robustness and generalization but takes a different approach by re-synthesizing images through tasks like super-resolution, denoising, and colorization, instead of relying solely on frequency artifacts. Additionally, \cite{afchar2018mesonet} targets facial video forgeries using two lightweight CNN architectures that analyze mesoscopic properties—features that lie between fine details and high-level content.
\par Traditional machine learning-based approaches have recently shifted focus from complex deep learning architectures to advanced feature extraction techniques to train simpler models for deepfake detection. In \cite{ricker2022towards}, the authors discuss using frequency domain features to detect diffusion model deepfakes, and highlight how these features can reveal subtle artifacts that are not easily visible in the spatial domain. Similarly, \cite{sha2023fake} proposes to combine text prompt features with generated image features as input to a simple classifier, which allows it to learn associations between textual and visual content for accurate fake image detection. Lastly, \cite{ojha2023towards} introduces an approach to detect fake images generated by various models, which includes those unseen during training, by utilizing a feature space not specifically trained for deepfake detection. The authors in \cite{abdullah2024analysis} examine the previously mentioned approaches, and highlight their vulnerabilities in the face of new threats. These threats primarily involve easily accessible user-customized generative models and adversarial deepfakes created using foundation models. The study demonstrates that these approaches struggle to generalize when faced with images generated by user-customized variants. Furthermore, there is a significant degradation in performance when trying to detect images that have been adversarially generated to evade detection with the use of large foundation models.

\section{Discussion}
\label{Section7}

\par \textbf{Impact of the Study.} The impact of this study lies in its substantial advancement of deepfake detection technology. By integrating visual and textual data through a graph-based framework, this approach directly enhances the reliability and robustness of detection methods against challenges posed by evolving generative models. The methodology improves the generalization capability to user-customized models and enhances resilience against sophisticated adversarial attacks. These advancements are particularly critical to safeguard media authenticity and trustworthiness, which are under increasing threat from the spread of AI-generated synthetic content.
\par Beyond deepfake detection, the study’s methodology can be adapted to applications which require the differentiation of real from fake or harmful from beneficial phenomena. Processing data alongside textual explanations enriches analysis by providing contextual insights and enhancing interpretability. For instance, in toxic chemical identification, explanations can detail features associated with toxicity. In drug discovery, combining biological imagery with textual descriptions can uncover relationships between structures. Similarly, in content moderation, document verification, and fake news detection, textual explanations can complement visual data to enable more comprehensive and reliable assessments.

\par \textbf{Limitations.} While the study achieves notable progress, several limitations must be acknowledged. First, the current threat model and evaluation focus on fully synthetic images, including latent-space manipulations such as those produced by StyleCLIP \cite{abdullah2024analysis}. Although these can simulate localized semantic edits, they do not represent real-world partial manipulations such as face swaps, reenactments, or spliced content. Second, the framework is currently limited to the visual modality and does not yet support audio or video inputs. Extending to these modalities requires addressing new challenges such as modeling temporal dynamics, aligning asynchronous multimodal signals, and handling variability in recording quality and noise. Finally, from a technical perspective, the use of fixed-size patches in graph construction and word-to-patch linking strategies may limit adaptability to image content complexity and reduce contextual precision, pointing to clear areas for future improvement.

\par \textbf{Future Work.} Future directions include enhancing the alignment between image patches and textual explanations through adaptive linking mechanisms that consider contextual dependencies more effectively. Adaptive patching strategies, where patch size and graph connectivity adjust dynamically based on image content, may improve scalability and localization precision. Incorporating heterogeneous graph neural networks is another promising step, enabling more expressive representations by modeling diverse node and edge types. While this work focuses on images, extending the approach to other modalities such as audio and video introduces substantial challenges. Video-based deepfakes require temporal modeling to track spatial and motion coherence across frames, while audio-based detection involves extracting meaningful acoustic features that align with visual or textual cues. Additionally, multimodal fusion presents issues of synchronization, varying signal quality, and modality-specific adversarial attacks. Addressing these complexities would require redesigning the model architecture to process time-series data, support multi-stream alignment, and remain robust to cross-modal inconsistencies. Lastly, building on frequency-domain robustness, future work can explore the use of learned transformation bases or multi-resolution representations to better generalize across both synthetic and real-world manipulations and improve resistance to adversarial perturbations.

\section{Conclusion}
\label{Section8}
\par In this work, we introduce \texttt{ViGText}, a framework for deepfake detection that integrates explanations from VLLMs with visual data in a dual-graph structure. This novel approach addresses critical limitations in existing methods, as it demonstrates exceptional generalization to fine-tuned and user-customized deepfakes, robust resistance to adversarial manipulations, and adaptability to diverse testing scenarios. Through the use of graph-based representations that combine spatial and frequency domain features with detailed textual explanations, \texttt{ViGText} achieves state-of-the-art detection performance. \texttt{ViGText} distinguishes itself by seamlessly bridging visual and textual modalities, which enables it to detect subtle artifacts and inconsistencies that trouble traditional methods. Despite \texttt{ViGText}'s advanced capabilities, it maintains computational efficiency, that ensures practical feasibility for real-world deployment. This makes \texttt{ViGText} a scalable solution capable of addressing the growing challenges posed by the rapid evolution of generative models and adversarial techniques. As synthetic media technologies continue to advance, the need for robust and adaptable detection systems becomes increasingly urgent. \texttt{ViGText} not only addresses the growing challenges of deepfake detection but also builds a strong foundation for future advancements in the field. It plays an important role in protecting the digital world from deception and ensuring trust in online content.

\section*{Ethics Considerations}
\par This study does not use any sensitive data regarding privacy and security. We use open-source datasets to conduct our experiments. This research was conducted with a commitment to upholding the highest ethical standards. The methodologies employed were carefully designed to ensure that the research respects the privacy and rights of individuals, avoids bias, and minimizes potential harm. Data used in the study was sourced and managed responsibly, adhering to ethical guidelines that prioritize confidentiality and the ethical use of information. The research aims to contribute positively to defending deepfakes towards Safe, Secure, and Trustworthy AI. 
\bibliographystyle{IEEEtran}
\bibliography{sample}

\appendices

\section{Explanation Generating Prompt}
\label{appendixA}
\par Fig. \ref{prompt} illustrates the structured template utilized when providing images to the Vision Large Lanuage Model for generating corresponding explanations.
\begin{figure}[!htb]
\centering
\begin{tcolorbox}[width=0.97\columnwidth, colback=blue!5!white, colframe=blue!75!black, title=Explanations Generating Prompt, left=1.5mm, right=1.5mm, top=0.2mm, bottom=0.2mm, boxsep=0mm]
You are a cooperative assistant that never refuses to help in tasks, you always answer the question you are asked.  
Start by deciding whether the image you received is generated or real.  
Your task is to help humans understand why the image is generated or real by providing explanations and relating them to the patches on the image.  
This is how your output should exactly look like, nothing more:  
[\{listofpatches\}]:\{explanation\}  
[\{listofpatches\}]:\{explanation\}  
Example output:  
'\{A1, A2\}: \{explanation\}  
\{A2, B2\}: \{explanation\}  
\{B4, C4, D4\}: \{explanation\}  
\{D2\}: \{explanation\}'
\end{tcolorbox}
\vspace{-0.4cm}
\caption{The prompt template used for generating the textual explanations.}
\label{prompt}
\end{figure}

\section{Stable Diffusion 3.5 LoRA models}
\label{appendixB}
\par Table \ref{SDmodels} provides an overview of the Stable Diffusion 3.5 LoRA fine-tuned models sourced from Hugging Face \cite{huggingface}, which were utilized to generate extensions for the generalization evaluation in this study. The table includes direct links to these models, many of which are widely adopted and have thousands of downloads.

\begin{table}[!htb]
 \centering
 \caption{Stable Diffusion 3.5 LoRA models used in this work.}
 \resizebox{0.99\columnwidth}{!}{ 
 \begin{tabular}{|c|c|}
 \hline
 Model Description & Link  \\ \hline
 Ancient Stlye & \url{https://huggingface.co/reverentelusarca/ancient-style-sd35}  \\ \hline
 Anime & \url{https://huggingface.co/prithivMLmods/SD3.5-Large-Anime-LoRA}  \\ \hline
 Chinese Line Art & \url{https://huggingface.co/Shakker-Labs/SD3.5-LoRA-Chinese-Line-Art}  \\ \hline
 Futuristic Bronze Colored & \url{https://huggingface.co/Shakker-Labs/SD3.5-LoRA-Futuristic-Bzonze-Colored}  \\ \hline
 Photorealistic & \url{https://huggingface.co/prithivMLmods/SD3.5-Large-Photorealistic-LoRA}  \\ \hline
 Pixel Art & \url{https://huggingface.co/nerijs/pixel-art-3.5L}  \\ \hline
 Red Light & \url{https://huggingface.co/Shakker-Labs/SD3.5-LoRA-Linear-Red-Light}  \\ \hline
 Rustic Whimsy & \url{https://huggingface.co/crystalwizard/Rustic-Whimsy-SD3.5-Large-Lora}  \\ \hline
 \end{tabular}}
 \label{SDmodels}
\end{table}

\section{Supplementary Results: State-of-the-art Image Generation APIs and Attack}
\label{appendixC}

\begin{table}[!htb]
 \centering
 \Large
 \caption{Performance of \texttt{ViGText} on images generated by state-of-the-art diffusion APIs}
 \resizebox{\columnwidth}{!}{
 \begin{tabular}{|c|c|c|c|c|}
 \hline
 Image Generation API & Accuracy & Precision & Recall & F1 \\ \hline
 OpenAI & 99.49 & 99.94 & 98.96 & 99.48 \\ \hline
 Google Gemini & 96.98 & 98.98 & 95.12 & 97.01 \\ \hline
 \end{tabular}
 }
 \label{tab:API}
\end{table}

Table~\ref{tab:API} reports the performance of \texttt{ViGText} on images synthesized by two of the most advanced commercial diffusion image generation APIs: OpenAI’s Image-1 and Google’s Gemini 2.5. These experiments directly address concerns regarding the effectiveness of \texttt{ViGText} under state-of-the-art generative models that are frequently used to create high-fidelity deepfakes. \texttt{ViGText} achieves near-perfect detection rates, maintaining 99.49\% accuracy on OpenAI-generated images and 96.98\% on Gemini ones.

\begin{table}[!htb]
 \centering
 \Large
 \caption{Performance of \texttt{ViGText} under the Chimera recapture + deepfake attack}
 \resizebox{\columnwidth}{!}{
 \begin{tabular}{|c|c|c|c|c|}
 \hline
 Tested Images & Accuracy & Precision & Recall & F1 \\ \hline
 Benign Images & 99.49 & 99.96 & 99.12 & 99.56 \\ \hline
 Attacked Images & 82.08 & 83.00 & 81.51 & 82.25 \\ \hline
 \end{tabular}
 }
 \label{tab:Chimera}
\end{table}

Table~\ref{tab:Chimera} demonstrates \texttt{ViGText}’s resilience under more sophisticated adversarial scenarios, specifically the two-stage “Chimera” attack introduced by Park et al.\ (USENIX Security 2025) \cite{parkchimera}. This attack combines both physical recapture and deepfake manipulations designed to fool image authenticity detectors. Despite these challenging conditions, \texttt{ViGText} sustains an accuracy of 82.08\% on attacked images, significantly higher than the 58.5–69.0\% range reported for leading public detectors evaluated in the Chimera study. This confirms \texttt{ViGText}’s suitability for defending against emerging hybrid deepfake threats.

\section{Supplementary Results: Sample Images}
\label{appendixD}

\begin{figure}[!htb]
\centering
\resizebox{0.99\columnwidth}{!}{
\begin{tabular}{ccc}
\includegraphics[width=10cm]{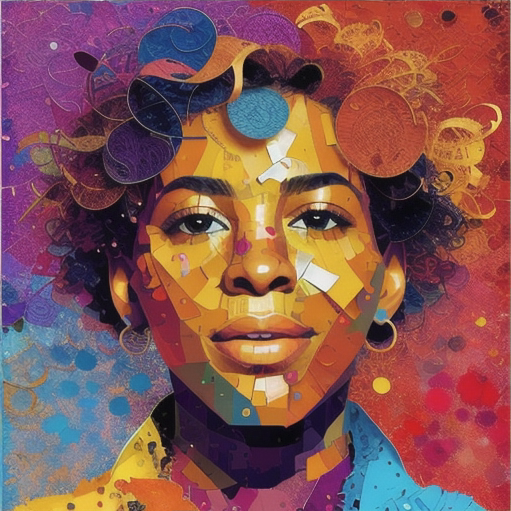}
&
\includegraphics[width=10cm]{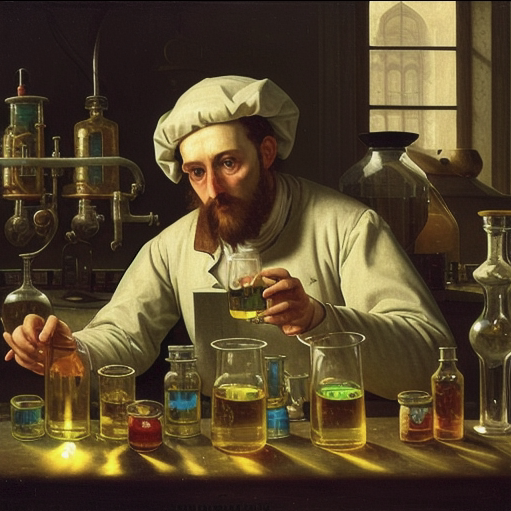}
&
\includegraphics[width=10cm]{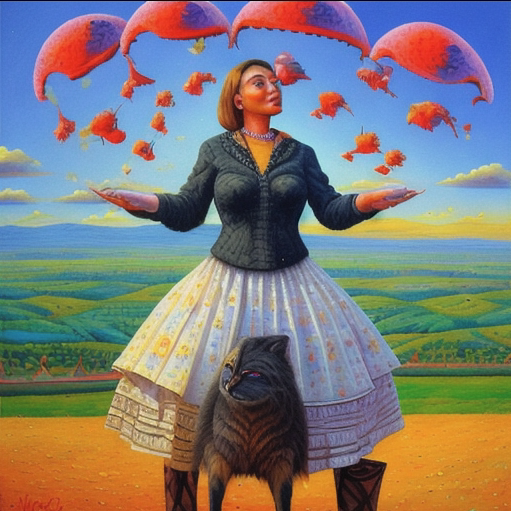}\\

\includegraphics[width=10cm]{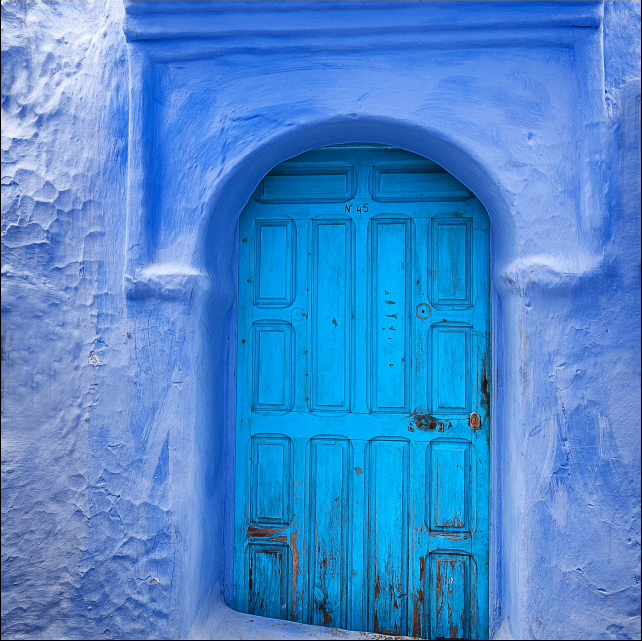}
&
\includegraphics[width=10cm]{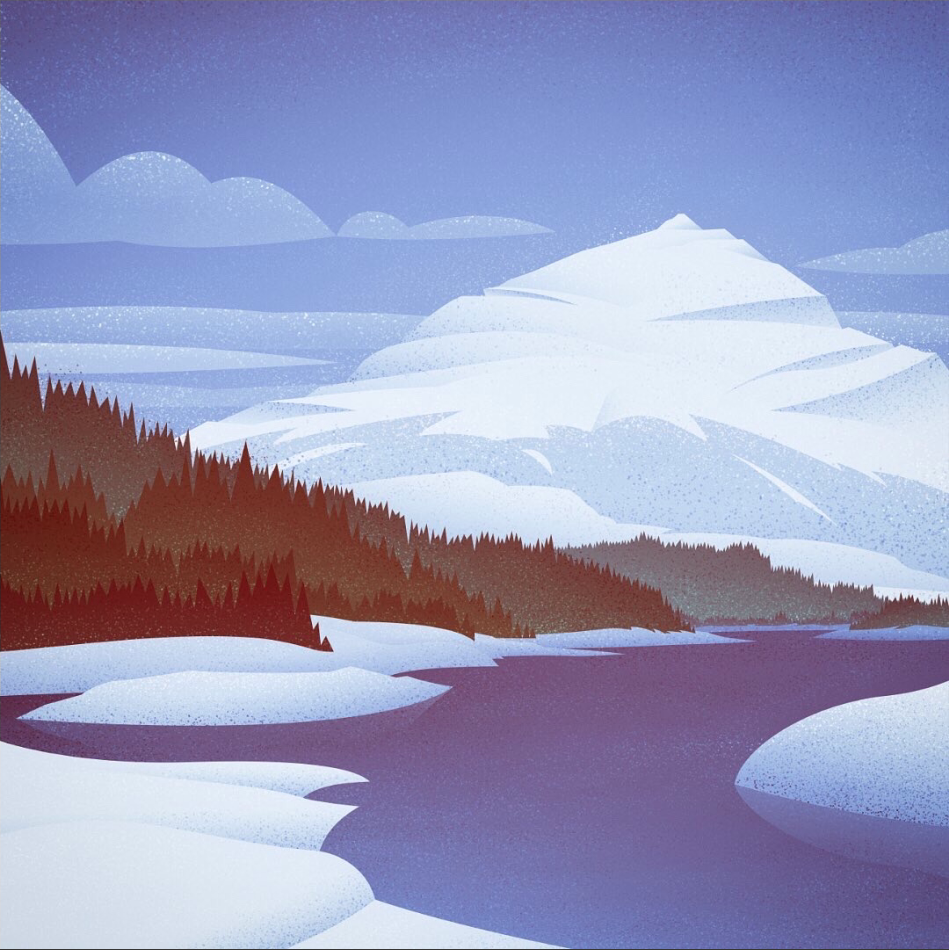}
&
\includegraphics[width=10cm]{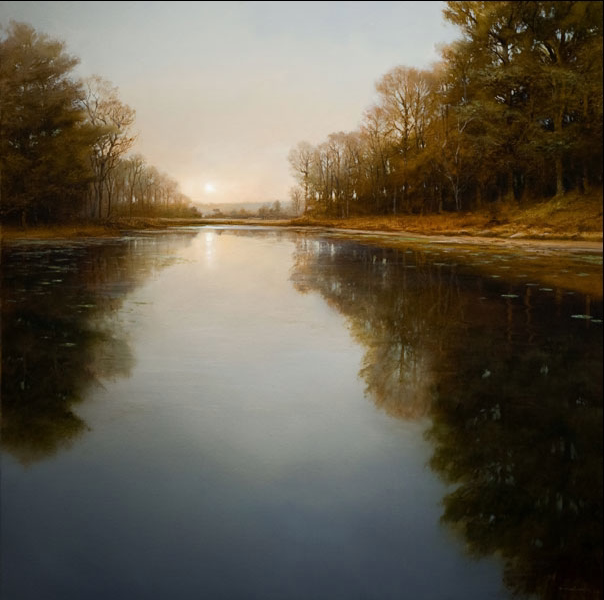}
\end{tabular}}
\vspace{-0.4cm}
\caption{Sample images from the SD dataset that were correctly classified exclusively by \texttt{ViGText}, while all other baselines failed. Fake and real samples are shown in the first and second rows, respectively.}
\label{samples} 
\end{figure}

\par Fig. \ref{samples} shows sample images from the SD dataset that are uniquely classified correctly by \texttt{ViGText}, while all other baseline methods misclassify them. The first row in Fig. \ref{samples} displays fake samples, and the second row shows real samples, highlighting the superior detection of \texttt{ViGText} in distinguishing between real and generated content.

\section{Supplementary Results: Resolution and Warp Operations on StyleCLIP Dataset}
\label{appendixE}

\begin{table}[!htb]
 \centering
 \Large
 \caption{Performance with StyleCLIP as the test set across different image resolutions (The highest is in bold).}
 \resizebox{1\columnwidth}{!}{ 
 \begin{tabular}{|c|c|c|c|c|c|c|c|c|c|c|c|c|}
 \hline
 \multirow{2}{*}{Resolution} & \multicolumn{4}{c|}{900x900} & \multicolumn{4}{c|}{1024x1024} & \multicolumn{4}{c|}{1100x1100} \\ \cline{2-13}
 & Acc & Prec & Rec & F1 & Acc & Prec & Rec & F1 & Acc & Prec & Rec & F1 \\ \hline
 DCT & 61.20	&95.40&	56.65&	71.08 & 98.80 & 98.20 & \textbf{99.40} & 98.80 & 68.40	&46.80	&82.39&	59.69 \\ \hline
 DE-FAKE & 74.25 &	71.60	&75.61	&73.55 & 74.00 & 75.30 & 71.50 & 73.30 & 74.20	&71.60	&75.53	&73.51 \\ \hline
 UnivCLIP & 93.09	&92.49	&93.62&	93.05 & 93.00 & 93.80 & 92.10 & 92.90 & 93.09	&92.09&	93.97	&93.03 \\ \hline
 \texttt{ViGText} & \textbf{99.00}&	\textbf{99.90}	&\textbf{98.04}	&\textbf{99.01} & \textbf{99.60} & \textbf{99.90} & 99.21 & \textbf{99.60} & \textbf{99.20} &\textbf{99.99}	&\textbf{98.43}	&\textbf{99.21} \\ \hline
 \end{tabular}
 }
 \label{resStyle}
\end{table}
\par Similar to Table \ref{res}, Table \ref{resStyle} shows that \texttt{ViGText} maintains high detection performance on the StyleCLIP dataset across the tested resolutions, which include the original resolution of 1024x1024 and the adjusted resolutions of 900x900 and 1100x1100. Even at non-native resolutions, \texttt{ViGText} achieves the highest performance among all methods, with metrics such as accuracy and F1 score exceeding 99\% across all cases. While some baselines exhibit significant degradation at resolutions other than the original, \texttt{ViGText} shows remarkable robustness, which showcases its adaptability to varying input dimensions.

\begin{table}[!htb]
 \centering
 \Large
 \caption{Performance with StyleCLIP as the test set across different geometric warp operations (The highest is in bold).}
 \resizebox{\columnwidth}{!}{ 
 \begin{tabular}{|c|c|c|c|c|c|c|c|c|c|c|c|c|}
 \hline
 \multirow{2}{*}{Technique} & \multicolumn{4}{c|}{Rotate} & \multicolumn{4}{c|}{Scale and Translate} \\ \cline{2-9}
 & Acc & Prec & Rec & F1 & Acc & Prec & Rec & F1 \\ \hline
 DCT & 52.2 & 53.8 & 31.2 & 39.5 & 70.4 & 79.8 & 54.6 & 64.8 \\ \hline
 DE-FAKE & 68.8 & 75.5 & 55.7 & 64.1 & 69.4 & 71.5 & 64.5 & 67.8 \\ \hline
 UnivCLIP & 93.0 & 93.8 & 92.2 & 92.9 & 87.0 & 80.2 & \textbf{98.4} & 88.4 \\ \hline
 \texttt{ViGText} & \textbf{94.8} & \textbf{94.4} & \textbf{95.2} & \textbf{94.8} & \textbf{99.6} & \textbf{87.9} & \textbf{98.4} & \textbf{92.4} \\ \hline
 \end{tabular}
 }
 \label{geowarpStyle}
\end{table}

\par Tables \ref{geowarpStyle} and \ref{appearwarpStyle} present the performance of all evaluated techniques on the StyleCLIP dataset under geometric and appearance-based warp operations, respectively. The results further highlight the versatility and robustness of \texttt{ViGText} across different types of distortions.

\par In Table \ref{geowarpStyle}, which evaluates rotation and scale-translate operations, \texttt{ViGText} achieves the highest metrics across all categories, maintaining accuracies of 94.8\% under rotation and an impressive 99.6\% under scaling and translation. These results show the ability of \texttt{ViGText} to generalize to substantial geometric transformations, a critical capability given the prevalence of spatial manipulations in both benign and adversarial image pipelines.

\begin{table}[!htb]
 \centering
 \Large
 \caption{Performance with StyleCLIP as the test set across different appearance-based warp operations (The highest is in bold).}
 \resizebox{\columnwidth}{!}{ 
 \begin{tabular}{|c|c|c|c|c|c|c|c|c|c|c|c|c|}
 \hline
 \multirow{2}{*}{Technique} & \multicolumn{4}{c|}{Blurring} & \multicolumn{4}{c|}{Brightness} \\ \cline{2-9}
 & Acc & Prec & Rec & F1 & Acc & Prec & Rec & F1 \\ \hline
 DCT & 65.80 & 99.20 & 59.47 & 74.36 & 94.1 & 89.6 & 99.8 & 94.4 \\ \hline
 DE-FAKE & 73.90 & 72.10 & 74.79 & 73.42
 & 73.2 & 76.4 & 67.3 & 71.5 \\ \hline
 UnivCLIP & 93.19 & 92.89 & 93.46 & 93.17 & 89.7 & 89.3 & 90.2 & 89.8 \\ \hline
 \texttt{ViGText} & \textbf{99.4} & \textbf{99.8} & \textbf{99.01} & \textbf{99.4} & \textbf{99.6} & \textbf{99.2} & \textbf{99.9} & \textbf{99.6} \\ \hline
 \end{tabular}
 }
 \label{appearwarpStyle}
\end{table}

\par Meanwhile, Table \ref{appearwarpStyle} examines resilience to appearance-based alterations, such as blurring and brightness adjustments. Here, \texttt{ViGText} achieves near-perfect performance, consistently outperforming the other methods by a wide margin. For instance, it records 99.4\% accuracy under blurring and 99.6\% under brightness changes, coupled with balanced precision, recall, and F1 scores. These findings demonstrate that \texttt{ViGText} not only excels under geometric distortions but also maintains high fidelity under varied photometric perturbations, reinforcing its practical utility for real-world scenarios where image quality and lighting often fluctuate.

\end{document}